\documentclass[runningheads]{llncs}

\usepackage{eccv}
\usepackage{eccvabbrv}
\usepackage{graphicx}
\usepackage{booktabs}
\usepackage[accsupp]{axessibility}  %
\usepackage[pagebackref,breaklinks,colorlinks,citecolor=eccvblue]{hyperref}
\usepackage{orcidlink}

\PassOptionsToPackage{capitalize}{cleveref}

\usepackage{microtype}

\usepackage{graphicx}
\usepackage[table]{xcolor}
\usepackage{tikz}
\usepackage{wrapfig}
\usepackage{adjustbox}
\usepackage{lscape}

\usepackage{booktabs}
\usepackage{tabularx}
\usepackage{multirow}
\usepackage{multicol}
\usepackage{longtable}
\usepackage{makecell}

\usepackage{amsmath}
\usepackage{amssymb}
\usepackage{amsfonts}
\usepackage{mathtools}
\usepackage{bm}
\usepackage{nicefrac}
\usepackage{dsfont}

\usepackage[ruled,vlined]{algorithm2e}

\usepackage{enumitem}
\usepackage{ragged2e}
\usepackage{cuted}
\usepackage{etoolbox}
\usepackage{xspace}
\usepackage{pifont}

\usepackage[hypcap=true]{caption}
\usepackage{subcaption}

\usepackage{url}
\usepackage{cleveref}

\usepackage{fontawesome5}

\usepackage{orcidlink}

\crefname{equation}{Eq.}{Eqs.}
\Crefname{equation}{Equation}{Equations}
\crefname{algorithm}{Alg.}{Algs.}
\Crefname{algorithm}{Algorithm}{Algorithms}
\crefname{section}{Sec.}{Secs.}
\Crefname{section}{Section}{Sections}
\crefname{figure}{Fig.}{Figs.}
\Crefname{figure}{Figure}{Figures}
\crefname{table}{Tab.}{Tabs.}
\Crefname{table}{Table}{Tables}

\definecolor{bblue}{RGB}{33,102,172}

\newcommand{\plusmark}{\textcolor{teal}{$\bm{+}$}}

\begin{document}

\title{Vitality-Aware Compression for Efficient Image-to-Shape Diffusion Transformers} 

\titlerunning{Vitality-Aware Compression for Image-to-Shape DiT}

\author{
Jaeah Lee\inst{1}$^\star$\orcidlink{0009-0004-2648-8523}
\and
Hyunjin Kim\inst{1}$^\star$\orcidlink{0009-0003-0922-6401}
\and
Jaewoong Cho\inst{1}\orcidlink{0009-0007-5864-9514}
\and
Gihyun Kwon\inst{2}$^\dagger$\orcidlink{0000-0002-2398-5282}
}
\authorrunning{J.~Lee et al.}

\institute{KRAFTON AI, Republic of Korea
\and
Amazon, Australia
}

\def\thefootnote{$\star$}\footnotetext{Authors contributed equally to this work.}
\def\thefootnote{$\dagger$}\footnotetext{Worked done at KRAFTON AI.}

\maketitle

\begin{figure}
    \centering
    \includegraphics[trim={0mm 0mm 0mm 0mm}, clip, width=1.00\linewidth]{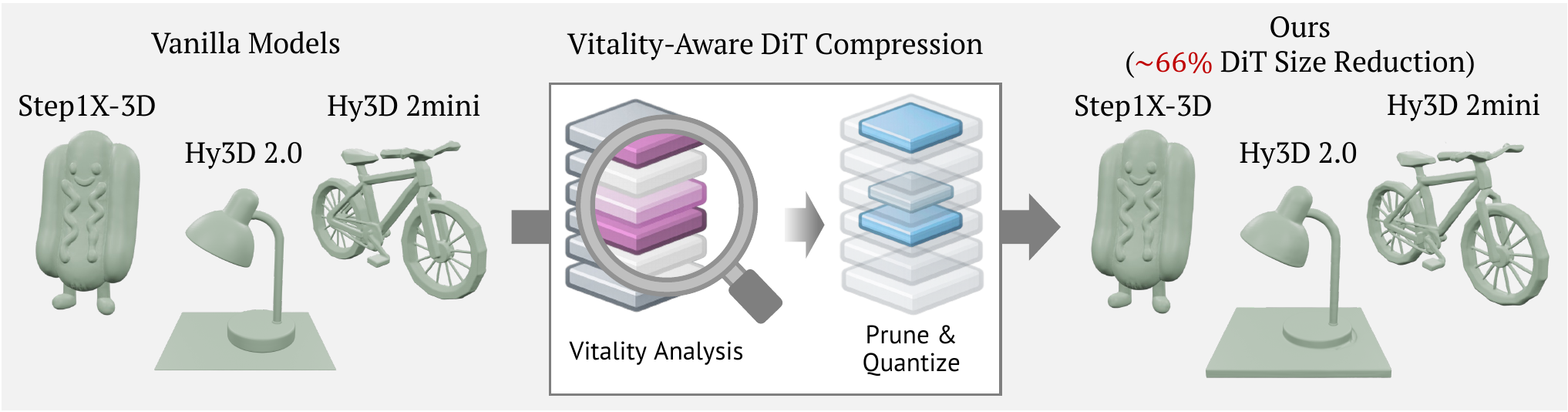}
    \captionof{figure}{
    In this paper, we introduce a vitality-guided Diffusion Transformer (DiT) compression pipeline for image-to-3D shape generation. Our approach reduces model size while preserving synthesis quality.
    }
    \label{fig:teaser}
\end{figure}

\begin{abstract}
We propose the \textbf{\textit{first}} compression approach for image-to-shape Diffusion Transformers (DiTs) that substantially reduces model size while preserving geometric fidelity.
Despite remarkable progress in 3D shape generation, large DiT-based models remain computationally prohibitive in resource-constrained settings.
Furthermore, it is difficult to directly transfer existing diffusion model compression strategies developed for different domains to 3D generation, and prior 3D efficiency approaches focus primarily on inference speed rather than backbone compression.
To address this limitation, we build a geometry-aware compression framework tailored to image-to-shape DiTs.
Guided by the observation that 3D DiT layers exhibit non-uniform importance for geometry synthesis, we introduce a vitality-guided framework integrating structured pruning, adaptive quantization, and targeted fine-tuning.
Our method achieves up to \textbf{66\%} model-size reduction across state-of-the-art image-to-3D models while maintaining synthesis fidelity comparable to full-sized counterparts.
This highlights the potential of our framework as a plug-and-play solution for efficient 3D shape generation across diverse models.
\keywords{Model Compression \and 3D Geometry Generation \and Image-to-3D Synthesis \and Diffusion Transformers \and Efficient Generative Models}
\end{abstract}

\section{Introduction}

The rapid expansion of 3D content across AR/VR, gaming, and embodied AI has intensified the demand for scalable and automated 3D shape generation.
Recent image-to-3D synthesis has progressed from GAN-based priors~\cite{henzler2019platonicgan,zhang2022meshinversion,yuan2023goae} and Large Reconstruction Models~\cite{hong2023lrm,gslrm2024,TripoSR2024} to 3D-native Diffusion Transformer (DiT) models~\cite{wu2024direct3d,zhang2024clay} and flow-matching frameworks~\cite{xiang2024structured,zhao2025hunyuan3d,li2025step1x}, achieving geometrically consistent meshes from a single image.
However, the DiT backbones in these pipelines often exceed 2.5\,GB in parameter size alone, limiting their use in real-time and resource-constrained environments.

\begin{figure*}[t]
    \centering
    \includegraphics[trim={0mm 3mm 9mm 0mm}, clip, width=1.00\linewidth]{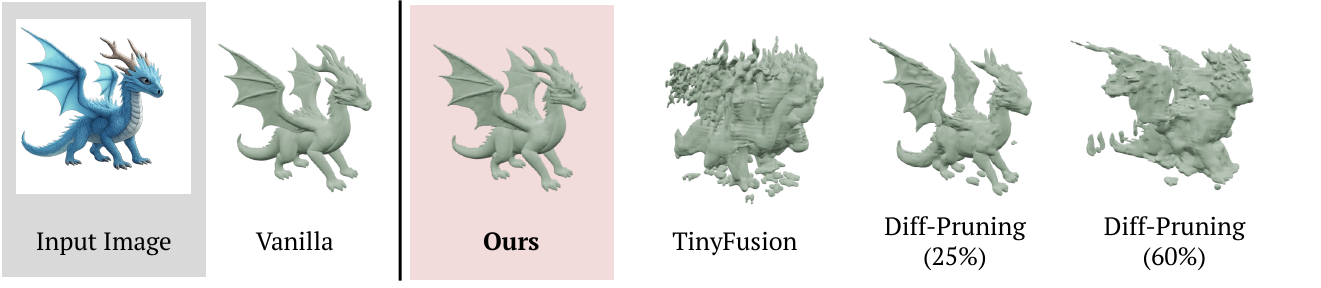}
    \captionof{figure}{
    \textbf{Domain Gap of DiT Compression on Step1X-3D.}
    Diffusion model compression strategies designed for image generation~\cite{fang2025tinyfusion, fang2023structural} do not directly transfer to image-to-shape synthesis, as preserving visual quality does not guarantee geometric fidelity.
    This highlights the need for an alternative approach to shape generation.
    }
    \label{fig:intro_problem}
\end{figure*}

While diffusion model compression has been actively studied for image~\cite{lee2024dit, fang2025tinyfusion, you2025layer, fang2023structural, chang2024sparsedit} and video~\cite{peruzzo2025adaptor, yin2025slow} generation, where redundancy mainly arises from spatial or temporal correlations, these strategies do not address the structural demands of 3D shape synthesis.
As shown in \cref{fig:intro_problem}, directly applying diffusion compression methods originally designed for image generation to shape generation results in severe geometric degradation, including structural collapse, distorted topology, and loss of fine details.
This discrepancy arises from the differences between 2D and 3D generation, as 3D models must maintain globally consistent geometry across viewpoints, and even small perturbations in the denoising process can propagate into structural artifacts~\cite{hong2023debiasing}.
Meanwhile, existing 3D efficiency approaches~\cite{TripoSR2024, lai2025unleashing} mainly focus on inference acceleration rather than backbone compression, providing limited benefit for memory-constrained applications.

Recent studies on DiT-based text-to-image~\cite{avrahami2025stable} and text-to-video~\cite{kim2025tv} synthesis have revealed that only a subset of layers significantly influence output quality.
However, these studies analyze per-layer importance for tasks such as perceptual editing or evaluating image-domain quality, rather than for preserving geometric fidelity under permanent structural compression.
To this end, we directly measure each layer's contribution to 3D synthesis quality via per-layer ablation with Earth Mover's Distance (EMD) on the generated point clouds.
This analysis reveals that 3D DiTs exhibit non-uniform layer importance, and that double-block and single-block layers display different sensitivity patterns, necessitating separate treatment during compression.

Building on these findings, we propose a simple yet effective compression pipeline: we first prune low-vitality layers using separate thresholds for double and single block layers to preserve both global coherence and local detail.
We then apply adaptive quantization, assigning higher precision to geometrically critical layers and more aggressive compression to less vital ones.
Finally, to mitigate performance degradation during compression, we adopt a targeted fine-tuning strategy that updates only the least-vital retained layers, resulting in efficient distillation.

To the best of our knowledge, our approach is the \textbf{\textit{first}} to systematically reduce both the parameter count and bit-width of image-to-shape DiTs and its application to physical model compression.
It achieves up to \textbf{66\%} model-size reduction while preserving synthesis quality across multiple state-of-the-art models, including Step1X-3D~\cite{li2025step1x} ($-$\textbf{65.63\%}), Hunyuan3D 2.0 ($-$\textbf{66.37\%}), and Hunyuan3D 2mini~\cite{zhao2025hunyuan3d} ($-$\textbf{44.50\%}).
This suggests potential as a plug-and-play solution for efficient 3D shape generation.

To summarize, our main contributions are:
\begin{itemize}[topsep=2pt]
    \item We present a per-layer analysis showing that 3D DiT layers exhibit non-uniform importance with distinct patterns across double- and single-block modules, and introduce an EMD-based vitality metric that directly measures each layer's contribution to 3D synthesis quality.
    \item We propose a simple yet effective compression pipeline that leverages per-layer vitality to guide structured pruning and adaptive mixed-precision quantization, using block-type-specific thresholds to preserve both global geometric coherence and fine surface detail.
    \item We introduce a selective fine-tuning strategy that updates only the least-vital retained layers in each module, improving distillation efficiency while recovering performance close to that of the original model.
    \item We validate our framework on three state-of-the-art DiT-based models, achieving up to 66\% model-size reduction with minimal degradation in synthesis quality.
\end{itemize}

\section{Related Work}

\paragraph{3D Generative Models.}
3D generative models have evolved across diverse representations, including voxels~\cite{wu20163dgan,xie2020genvoxelnet,mittal2022autosdf}, point clouds~\cite{luo2021dpm,zhou2021pvd,vahdat2022lion}, implicit fields~\cite{zheng2022sdfstylegan,hui2022neuralwavelet,shue2023triplanediffusion,chou2023diffusionsdf}, and meshes~\cite{nash2020polygen,siddiqui_meshgpt_2024}.
Early GAN-based approaches such as EG3D~\cite{chan2022efficient} and pi-GAN~\cite{chan2021pi} demonstrated view-consistent synthesis but were constrained by limited category diversity and training data. Diffusion-based models later improved geometric fidelity, with Shape-E~\cite{jun2023shap} introducing one of the first text-to-3D diffusion frameworks and inspiring methods that jointly model geometry and appearance.
More recently, large-scale datasets such as Objaverse~\cite{deitke2023objaverse} enabled Large Reconstruction Models (LRMs)~\cite{hong2024lrm,tang2024lgm,gslrm2024,TripoSR2024,liu2023one,xu2024instantmesh} for single-pass 3D synthesis, while systems including 3DTopia-XL~\cite{chen2025primx} and GaussianAnything~\cite{lan2024ga} leverage triplane and scalable Gaussian representations for open-domain generation.
To mitigate this, recent methods adopt two-stage pipelines that combine compact geometry generation with multi-view diffusion for texturing~\cite{zhang2024clay,li2025step1x,zhao2025hunyuan3d}, while others explore Structured Latent representations~\cite{xiang2024structured}.
Despite these advances, substantial memory and computational demands remain a key obstacle to the broader adoption of 3D generative modeling.

\paragraph{Model Compression for Transformer-based Models.}
While recent 3D generative models~\cite{li2025step1x,zhao2025hunyuan3d,xiang2024structured} have achieved remarkable improvements in fidelity, they still suffer from extremely high memory consumption.
Although methods like Turbo3D ~\cite{hu2025turbo3d} and FlashVDM ~\cite{lai2025unleashing} attempt to address efficiency, they mainly focus on accelerating inference rather than fundamental model compression.
In the broader Transformer literature, prior work have shown various pruning approaches, including attention head, block, and layer pruning ~\cite{fan2019reducing, lee2024dit, fang2025tinyfusion}, can effectively reduce model complexity while maintaining performance.
Extensive research have explored quantization, spanning from low-bit BERT models ~\cite{zafrir2019q8bert,shen2020q} to recent DiT-specific schemes ~\cite{wu2024ptq4dit, chen2025q, hwang2025tq}.
These methods consistently demonstrate that substantial memory savings can be achieved without compromising generation quality.
In addition, knowledge distillation techniques ~\cite{sanh2019distilbert, jiao2019tinybert, wang2020minilm} have proven effective in recovering accuracy after compression.
Despite these advances, 3D generative modeling lacks a systemic investigation into Transformer layer vitality and its application to pruning and quantization, which forms the central motivation of our work.

\section{Method}
\label{sec:03_method}

\begin{figure}[t]
    \centering
    \includegraphics[trim={1.5mm 2mm 0mm 0mm}, clip, width=1.0\linewidth]{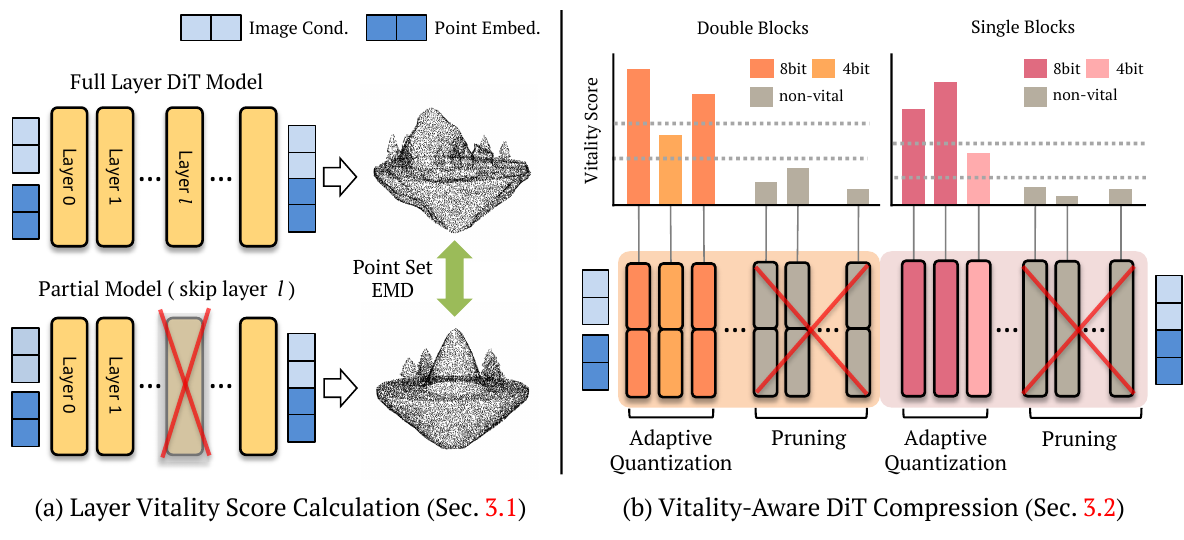}
    \caption{\textbf{Method Overview.}
    \textbf{(a)}
    We measure the contribution of each DiT layer $l$ by computing the point cloud distance between the full model output and the layer-ablated output using Earth Mover’s Distance (EMD).
    \textbf{(b)}
    Based on the vitality scores, we prune low-vitality layers using separate thresholds for double- and single-block layers. We then apply adaptive quantization, assigning 8-bit precision to highly vital layers and 4-bit to less vital ones.
    }
    \label{fig:method1}
\end{figure}

Our primary goal is to physically reduce the model size of 3D shape generation DiT architectures.
In the DiT backbone used for image-to-3D geometry synthesis, layers are organized into double block and single block modules.
In a double block module, modality-specific latent streams (\eg, noise and conditional tokens) are maintained separately while interacting through shared attention modules.
In contrast, single blocks operate on a unified latent representation after modality fusion.

To this end, we implement a structured compression pipeline that begins with per-layer contribution analysis (\cref{subsec:03-1_analysis}).
This allows us to identify redundant layer whose importance is negligible.
Subsequently, we apply structured pruning and layer-wise adaptive quantization guided by analyzed vitality,
constructing a lightweight model that almost preserves the performance of the original model.
Finally, we fine-tune the compressed model to closely match the accuracy of the full model (\cref{subsec:03-3_finetuning}).

\subsection{Layer-Wise Vitality Analysis for 3D Shape DiTs}
\label{subsec:03-1_analysis}

\begin{figure*}[t]
    \centering
    \includegraphics[trim={1.5mm 2mm 1mm 0mm}, clip, width=1.0\linewidth]{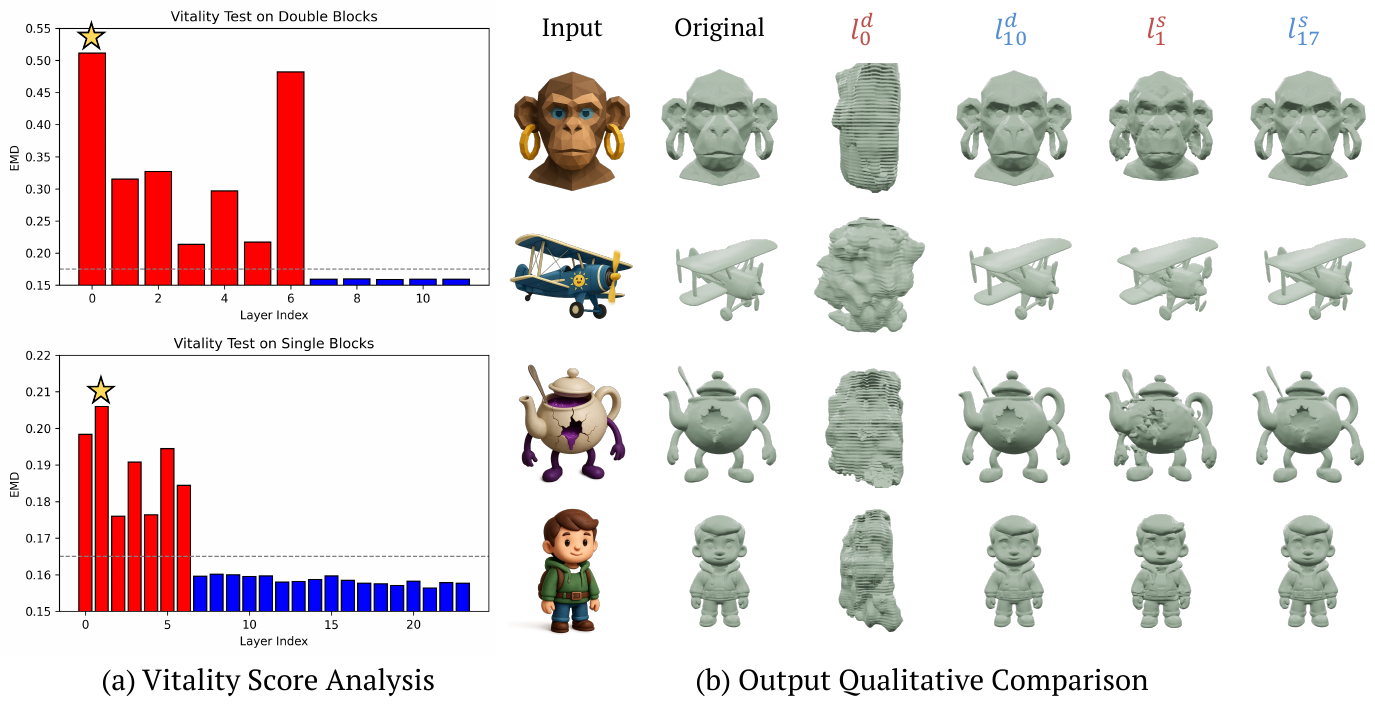}
    \caption{\textbf{Vitality Analysis Results on Step1X-3D.}
    \textbf{(a)}
    Layers marked in red are vital, contributing significantly to synthesis, while most layers appear in blue, indicating negligible contribution.
    The layer with the highest vitality in each module is marked with a star.
    \textbf{(b)}
    Removing vital layers leads to noticeable degradation in shape quality, while removing non-vital layers causes only minor differences.
    Here, $l_{i}^{d}$ denotes the $i$-th double-block layer and $l_{j}^{s}$ the $j$-th single-block layer, with indexing starting from 0.
    }
    \label{fig:analysis}
\end{figure*}

We begin by measuring the contribution of each layer in the 3D DiT model to the final output.
In prior work~\cite{avrahami2025stable} on text-to-image (T2I) generative models, the vitality of a layer is evaluated by comparing the outputs of the full DiT framework with that of a model where a target single layer $l$ is removed.
The perceptual difference between the two outputs is measured using the DINO~\cite{caron2021emerging} distance, and layers that induce larger discrepancies regarded as more important.

Following a similar principle, we analyze the Image-to-3D DiT layers using layer ablation in \cref{fig:method1}~(a).
Given the same conditional input image $y$, we generate a point set using the full model $\theta_{\text{full}}$ and layer-ablated model $\theta_{-l}$ by removing $l$-th layer.
The distance between these point sets then serves as a quantitative indicator of vitality.
Since the perceptual distance used in the image domain cannot be directly applied, here we require a metric suitable for 3D point sets. We therefore adopt Earth Mover's Distance (EMD) to measure the vitality of 3D DiT layers, as it effectively captures overall geometric differences between point sets.

For a conditional image $y$, our vitality score is defined as:
\begin{equation}
{\text{vitality}}(l)
= \mathbb{E}_{y \sim \mathcal{D}}
\left[
\min_{\Gamma \in \mathcal{P}_n} \;\frac{1}{n}\sum_{i=1}^{n}\sum_{j=1}^{n} 
\Gamma_{ij}\,\bigl\| q_{\theta_{\text{full}}}^{(i)}(y) - q_{\theta_{-l}}^{(j)}(y) \bigr\|_2
\right],
\end{equation}
where $\mathcal{D}$ is an image dataset, $n$ denotes the number of points in each point cloud, $q_{\theta_{\text{full}}}(y)$ is point cloud generated from full model,  $q_{\theta_{-l}}(y)$ is point cloud generated from layer $l$ removed model, and permutation matrices are defined as $\mathcal{P}_n 
= \Bigl\{ \Gamma \in \{0,1\}^{n \times n} \;\Big|\; 
\sum_{j=1}^{n} \Gamma_{ij} = 1, \;
\sum_{i=1}^{n} \Gamma_{ij} = 1, \;
\forall i,j
\Bigr\}$.

Unlike Chamfer Distance, which relies on nearest neighbor correspondences and mainly reflects local surface accuracy, EMD computes the optimal transport cost between two point sets, producing a one-to-one correspondence that captures the overall shape distribution.
This enables EMD to detect global structural distortions such as shifts, asymmetry, or large-scale misalignment that may arise when a layer responsible for geometric coherence is removed.
Moreover, EMD is less biased toward dense or unevenly sampled regions, ensuring consistent vitality evaluation across shapes of varying mesh density.
We provide a quantitative robustness comparison and the corresponding Chamfer Distance analysis in the appendix.

\Cref{fig:analysis}~(a) shows the results of our analysis on the Step1X-3D~\cite{li2025step1x} model, computed from 210 randomly generated images by DALL$\cdot$E~3~\cite{betker2023improving} using text prompts from Objaverse~\cite{deitke2023objaverse}.
Most layers are found to have vitality scores that converge close to zero, indicating negligible importance.
This pattern is consistent across both single and double block layers.
Similar trends are observed in other image-to-3D generation models, including Hunyuan3D 2.0 and Hunyuan3D 2mini~\cite{zhao2025hunyuan3d}, though with slightly weaker magnitudes.
Details of the analysis are provided in the appendix.

The qualitative analysis in \cref{fig:analysis}~(b) makes this effect more tangible.
Skipping vital double-block layers produces severe geometric distortions, such as unintended rotations, while removing vital single-block layers leads to degraded finer details and artifacts.
Conversely, omitting low-vitality layers in either cases barely effects the output.

\subsection{DiT Compression using Vital Layers}
\label{subsec:03-2_compression}

\subsubsection{Layer Pruning}
Based on the vitality scores, we determine which layers to prune using a threshold $\tau$.
Layers with vitality scores exceeding $\tau$ are classified as vital and retained, while the rest are pruned.
However, we observe that applying a single threshold across both double- and single-block layers causes performance degradation.
To mitigate this, we introduce separate thresholds, $\tau_d$ and $\tau_s$, for double- and single-block layers, respectively.
To determine these thresholds, we progressively remove layers starting from the lowest vitality score and monitor the distance to the vanilla model output.
The threshold is chosen at the point where a sharp drop in quality occurs.
We provide the detailed selection process in the appendix.

\subsubsection{Adaptive Quantization}
After pruning, we further reduce the model size through quantization.
Here, we also leverage the vitality scores to assign different bit-widths to each layer.
To minimize performance loss while maximizing compression, we define two groups: highly vital layers are quantized to 8-bit, and less-vital layers to 4-bit.
Similar to pruning, distinct thresholds are applied to double-block and single-block layers to avoid performance drops.
Since our method primarily focuses on layer-wise analysis, we apply weight-only quantization and do not consider activations.

\subsection{Distillation Fine-tuning}
\label{subsec:03-3_finetuning}

\begin{figure*}[t]
    \centering
    \includegraphics[trim={4.5mm 2mm 1mm 0mm}, clip, width=1.0\linewidth]{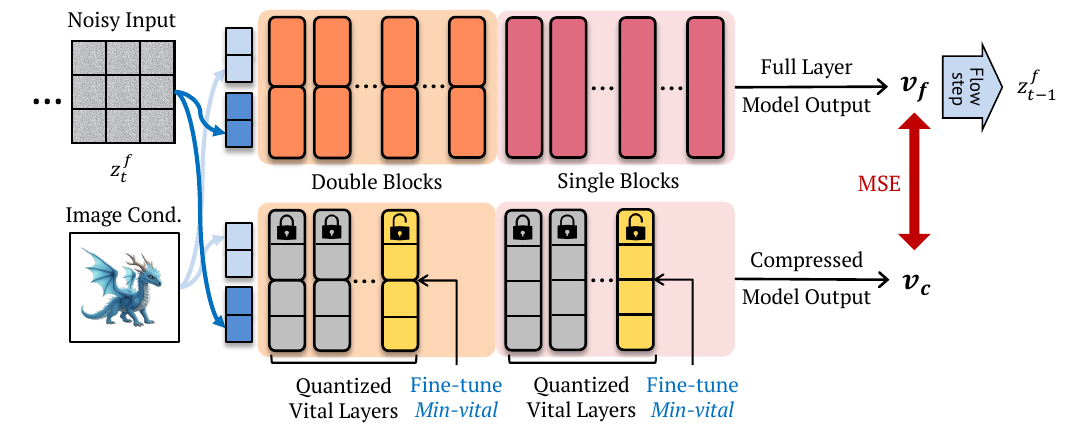}
    \caption{\textbf{Targeted Fine-tuning Pipeline.}
    To refine the compressed model, we fine-tune the minimally vital (\textit{Min-vital}) layer so that its output matches the full model.
    Specifically, along the full-model flow sampling path, we optimize the compressed student to reproduce the teacher’s output under the same conditions and latent input.
    }
    \label{fig:method2}
\end{figure*}

While our proposed pruning and quantization strategy yield an efficient compression, the resulting model may not entirely reproduce the behavior of the full model.
To bridge this gap, we perform fine-tuning so that the compressed model better follows the dynamics of the full model as shown in \cref{fig:method2}.
Unlike standard flow matching training, our approach focuses on maximizing similarity between the compressed and full models.
Specifically, we design a loss function to encourage the student to imitate the ODE path of the full model such as:
\begin{equation}
\mathcal{L}_{\text{Distill}}(\theta_c)
= \tfrac{1}{2} \, \big\| v^c(z^f_t, t, y) - v^f(z^f_t, t, y) \big\|_2^2
+ \tfrac{1}{2} \, \big\| v^c(z^f_t, t, \varnothing) - v^f(z^f_t, t, \varnothing) \big\|_2^2,
\end{equation}
where $v^c$ is model prediction output from compressed model $\theta_c$, $v^f$ is output from full model $\theta_{full}$, $z^f_t$ is latent of timestep $t$ sampled from full model, $y$ is input image condition, and $\varnothing$ is null condition.
In order to obtain more accurate distillation, we calculate distances for both of conditional and unconditional model predictions.
For each individual timestep, we optimize the parameters of weights from compressed model.
After single optimization step at timestep $t$, we jump into next step $t-1$ using flow sampling with full-model prediction output. 

However, fine-tuning all remaining vital layers is computationally inefficient.
In some cases, it causes the compressed student model to diverge further from the full teacher model, leading to degraded performance.
To mitigate this, we propose a selective fine-tuning strategy.
Specifically, we choose the vital layer with the lowest vital score (denoted as ``Min-vital'' in \cref{fig:method2}) of each DiT block and fine-tune only their weights, thereby avoiding excessive modification of vital layers.

\section{Experiment}
\label{sec:04_experiment}

\paragraph{Experimental Details.}
To validate our proposed method, we conduct experiments on three image-to-shape generation models. We use the state-of-the-art models Step1X-3D~\cite{li2025step1x}, Hunyuan3D 2.0 and 2mini~\cite{zhao2025hunyuan3d}. 
As described in~\cref{subsec:03-2_compression}, based on results of the vitality analysis, we set the standard for eliminating redundant layers and for setting thresholds to determine 8-bit and 4-bit layers.
For example, for Step1X-3D, we apply $\tau_d=0.17$ for double-block layers and $\tau_s=0.165$ for single-block layers, and set thresholds of 0.25 and 0.185 for double-block and single-block layers to determine the 8-bit and 4-bit layers.
During the non-vital layer fine-tuning stage, we use rendered images from subset 10K of Objaverse~\cite{deitke2023objaverse} dataset.
For Step1X-3D, we train with a learning rate of $10^{-8}$, and for Hunyuan3D 2.0 and 2mini, we used $10^{-4}$.
In both cases, we conduct fine-tuning process for 30K iterations for Step1X-3D, and 20k iterations for Hunyuan3D models.
For sampling, we use timestep of 30 for Step1X-3D and 20 for Hunyuan3D models.
We provide more experimental details in the appendix.

\paragraph{Evaluation Metrics.}
For evaluation, we employ two embedding-based metrics that measure semantic correspondence between input images and generated 3D meshes: \textbf{Uni3D-I}~\cite{zhou2023uni3d} and \textbf{OpenShape-I}~\cite{liu2023openshape}.
Both models compute similarity in a joint image–3D embedding space, providing an objective measure of alignment quality.
We report results on 200 image–shape pairs sampled from Objaverse~\cite{deitke2023objaverse}.
For validation, we generate 200 images using DALL$\cdot$E 3~\cite{betker2023improving} from text prompts originally provided by Objaverse~\cite{deitke2023objaverse}.

In addition, we measure the model size, specifically the memory footprint of its parameters, along with inference TFLOPs, generation latency, and peak VRAM usage, to evaluate spatial and computational efficiency after compression.
We also evaluate geometric consistency during compression using the \textbf{volume IoU (V-IoU)} and \textbf{surface IoU (S-IoU)} scores with rigid alignment, which are provided in the appendix.

\paragraph{Baselines.}
We compare our method with a diverse set of 3D generation approaches, spanning feedforward, diffusion, and transformer-based paradigms:
\begin{itemize}[topsep=2pt]
\item \textbf{Splatter Image}~\cite{szymanowicz24splatter}: a diffusion-based model that progressively generates 3D from images, achieving higher realism but often struggling with fine-grained alignment.
\item \textbf{TripoSR}~\cite{TripoSR2024}: a fast feedforward model that directly predicts 3D shapes from images, designed for lightweight inference but with limited geometric fidelity.
\item \textbf{LGM}~\cite{tang2024lgm}: a Gaussian-based feedforward approach that produces compact 3D representations, prioritizing efficiency over detailed reconstruction.
\item \textbf{Craftsman3D}~\cite{li2024craftsman}: a transformer-based DiT model with strong mesh generation quality, though requiring large memory and computation.
\item \textbf{TRELLIS}~\cite{xiang2024structured}: another state-of-the-art DiT-based architecture that excels in generating structured 3D meshes, but comes with significant model size overhead.
\end{itemize}

In addition, we compare with representative diffusion model compression methods originally designed for different domains, including TinyFusion~\cite{fang2025tinyfusion} and Diff-Pruning~\cite{fang2023structural}.
These methods are applied to the same 3D DiT backbone for fair comparison.
Detailed results are provided in the appendix.

\subsection{Quantitative Results}

\begin{table}[t]
\centering
\scriptsize
\caption{
\textbf{Quantitative comparison with baselines.}
Our method maintains high synthesis performance under compression
compared to original frameworks and other 3D generative models. VRAM indicates the peak VRAM usage.
}
\resizebox{0.95\textwidth}{!}{%
\begin{tabular}{
  @{}
  >{\RaggedRight\arraybackslash}p{0.22\textwidth}
  >{\centering\arraybackslash}p{0.14\textwidth}
  >{\centering\arraybackslash}p{0.16\textwidth}
  >{\centering\arraybackslash}p{0.14\textwidth}
  >{\centering\arraybackslash}p{0.14\textwidth}
  >{\centering\arraybackslash}p{0.14\textwidth}
  >{\centering\arraybackslash}p{0.15\textwidth}
  @{}
}
\toprule
\multirow{2}{*}{\textbf{Models}} & \multicolumn{2}{c}{\textbf{Metrics}} & \multicolumn{4}{c}{\textbf{Overhead}} \\
& {Uni3D-I $\uparrow$} & {OpenShape-I $\uparrow$} & {Size (GB) $\downarrow$} & {TFLOPs $\downarrow$} & {Latency (s) $\downarrow$} & {VRAM (GB) $\downarrow$} \\
\midrule
Splatter Img & 0.1800 & 0.0681 & 0.661 & 0.166& 0.02& 0.693\\
TripoSR      & 0.2994 & 0.1313 & 0.622 & 0.961& 0.04& 1.825\\
LGM          & 0.2482 & 0.1108 & 0.800 & 0.794& 0.04& 1.127\\
Craftsman3D  & 0.3519 & 0.1455 & 2.322 & 98.08& 3.36& 1.119\\
TRELLIS      & 0.3442 & 0.1455 & 2.175 & 527.5& 10.81& 6.625\\
\midrule
Step1X-3D               & 0.3586 & 0.1480 & 2.452 & 290.68& 6.23& 2.718\\
\textbf{Step1X-3D+Ours} & \textbf{0.3580} & \textbf{0.1489} & \textbf{0.843} & \textbf{113.24}& \textbf{2.78}& \textbf{1.206}\\
\midrule
Hy3D 2.0                 & 0.3582 & 0.1487 & 2.704 & 370.03& 5.85& 2.463\\
\textbf{Hy3D 2.0+Ours}  & \textbf{0.3601} & \textbf{0.1491} & \textbf{0.909} & \textbf{292.98}& \textbf{3.90}& \textbf{1.761}\\
\midrule
Hy3D 2mini               & 0.3614 & 0.1490 & 1.042 & 59.59& 1.28& 1.224\\
\textbf{Hy3D 2mini+Ours} & \textbf{0.3608} & \textbf{0.1484} & \textbf{0.578} & \textbf{54.64}& \textbf{1.14}& \textbf{1.135}\\
\bottomrule
\end{tabular}
}
\label{tab:comparison}
\end{table}

In \cref{tab:comparison}, we provide the quantitative comparison results between our proposed lightweight model and other baselines.
As already shown in the previous part, we use same baseline methods including reference models of Step1X-3D , Hunyuan3D 2.0, and Hunyuan3D 2mini.
For fair comparison, we only calculate parameter size of backbone models (U-Net or Transformer), without considering subsidiary networks such as autoencoder and condition encoders.

Comparing with early  methods of Splatter Image, TripoSR and LGM, the mesh quality and perceptual scores are largely degraded comparing with our methods although they have relative small model size.
With recent models of Craftsman3D and TRELLIS, quantitative scores are higher than other baselines, however they still do not outperform our best model (Hunyuan3D 2mini + Ours), in terms of mesh generation quality and model size.

We also compare each reference full model with its compressed version. Against the corresponding full models, our method reduces the model size by
more than 50\% for Step1X-3D and Hunyuan3D 2.0, and further compresses the
already compact Hunyuan3D 2mini with negligible degradation. In addition to
model size, \cref{tab:comparison} reports inference TFLOPs, generation latency,
and peak VRAM usage, all of which are consistently reduced by our method across
the evaluated models. Overall, our method reduces both spatial and computational
costs while preserving synthesis quality.

\begin{wrapfigure}{r}{0.6\linewidth}
    \centering
    \includegraphics[width=\linewidth]{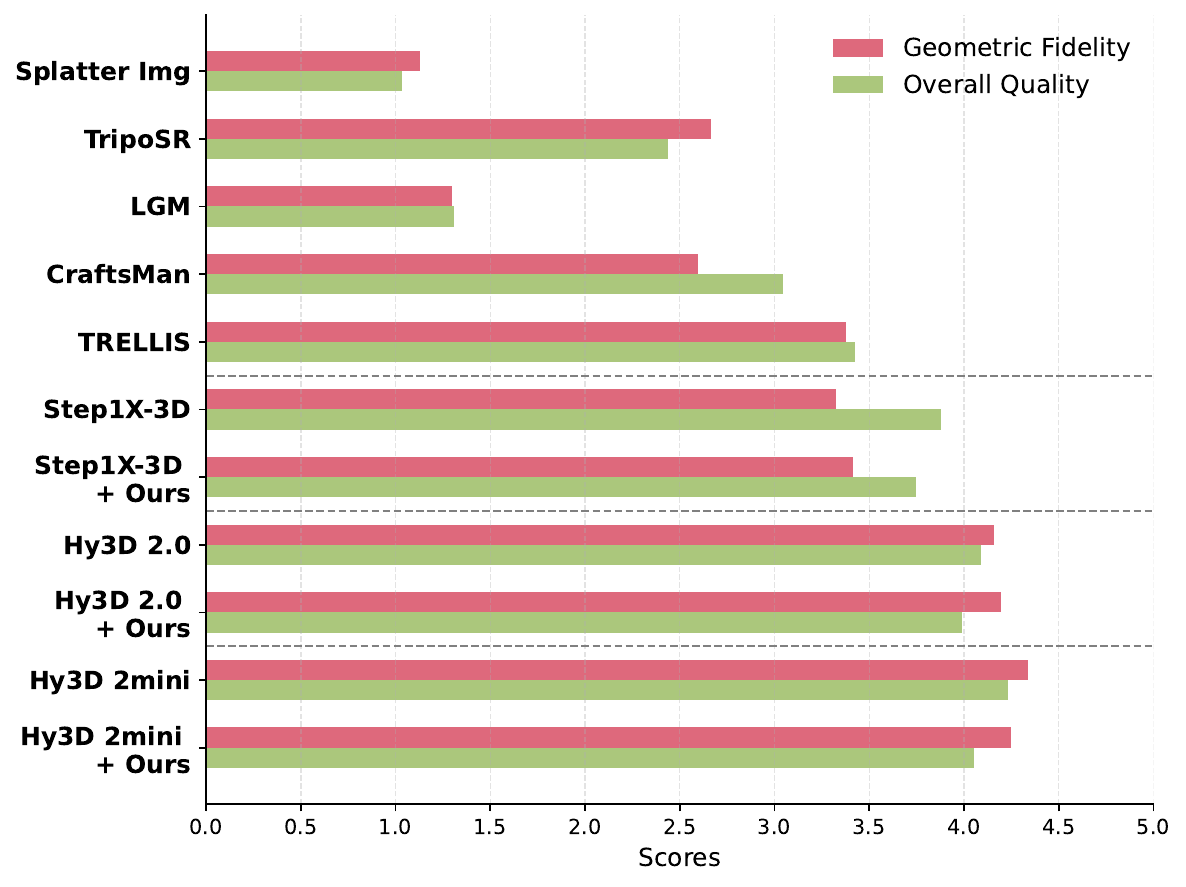}
    \caption{\textbf{User Study Results.}
    Our compression strategy preserves perceptual quality, achieving performance nearly indistinguishable from the full model.
    }
    \label{fig:user_study}
\end{wrapfigure}

To further access the perceptual quality of our proposed method, we present user study results in \cref{fig:user_study}. 
To evaluate the quality of 3D shape synthesis, participants were asked two questions: \textbf{1)} whether the correspondence between the image and the generated shape was reasonable (Geometric Fidelity), and \textbf{2)} whether the quality of the generated 3D mesh was satisfactory (Overall Quality).
Details of the experimental setup for the user study is provided in the appendix.

Consistent with our quantitative results, we observe that earlier works such as Splatter Image, LGM, and TripoSR exhibit substantially lower perceptual mesh quality compared to other models.
Recent methods, like Craftsman3D and TRELLIS, show improvements over the earlier models but still fall short of ours.
Notably, our compressed frameworks achieve high performance nearly indistinguishable from the full model baseline.
This also demonstrates that our compression method effectively preserves the performance of the full model.
\vspace{-10pt}  %

\subsection{Qualitative Results}
\begin{figure}
    \centering
    \includegraphics[trim={2mm 3mm 1mm 0mm}, clip, width=\linewidth]{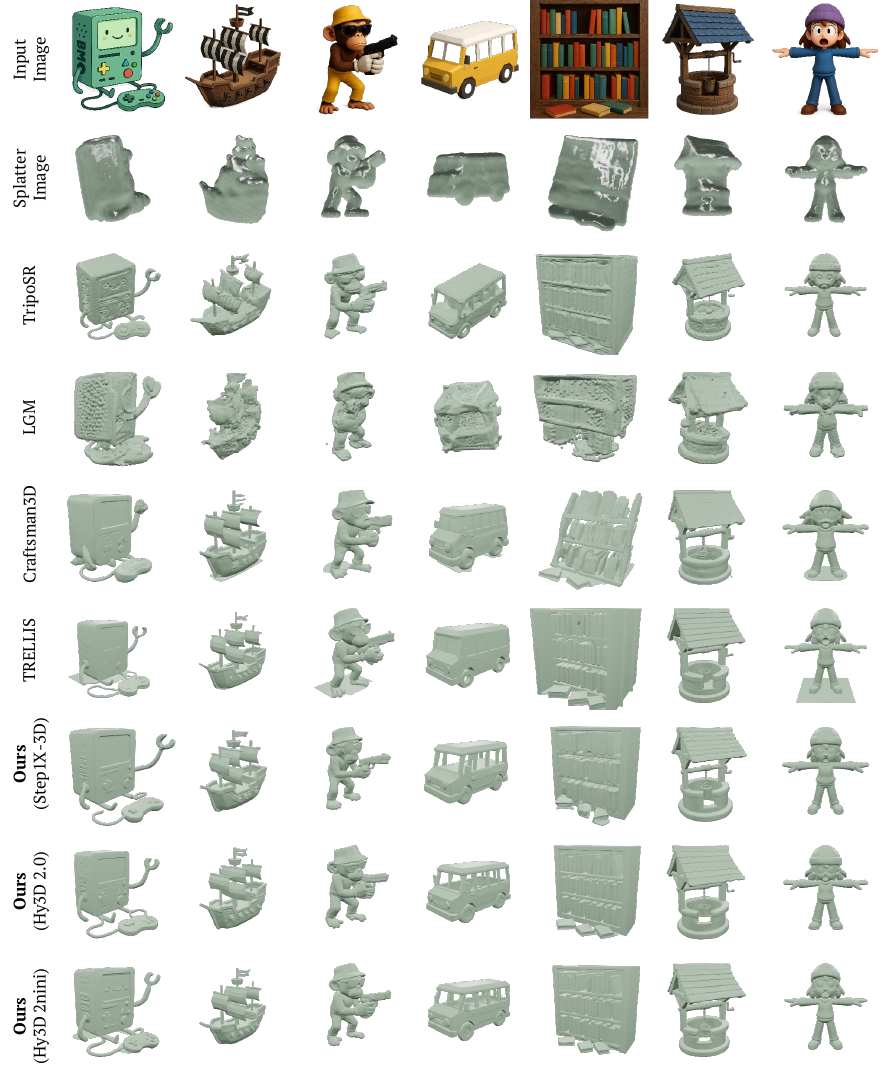}
    \caption{\textbf{Qualitative Comparison with Baselines.}
    For conditional image-to-3D mesh generation, earlier works such as Splatter Image, TripoSR, and LGM often produce meshes with lost details or struggle to match the alignment with the input image.
    Recent models like Craftsman3D and Trellis achieve good quality but still fall slightly short of ours in terms of fine details.
    Our models deliver superior perceptual performance while maintaining a significantly reduced model size compared to prior approaches.
    }
    \label{fig:qualitative_comparison}
\end{figure}

We qualitatively compare our method with representative baselines across different model families as shown in \cref{fig:qualitative_comparison}. 
Compared to the diffusion-based Splatter Image, which often struggle to capture fine details or maintain strong alignment with the input image, our approach achieves superior shape generation quality with smaller model sizes. 
Against feedforward models such as TripoSR and LGM, our method produces more detailed and faithful reconstructions, whereas the baseline often fails to capture fine image-specific features and exhibits artifacts.
In addition, compared to recent DiT-based models (Craftsman3D, TRELLIS), our framework generates meshes with sharper details and stronger image–shape correspondence.

\subsection{Ablation Study}

\begin{table*}[t]
\centering
\scriptsize
\caption{
\textbf{Quantitative Comparison on Ablation Study.}
We evaluate image--3D shape correspondence under various ablations
for Step1X-3D and Hunyuan3D models.
(\textbf{Bold}: best,\;
\underline{Underline}: second best,\;
\colorbox{yellow}{Colored}: within 1\% of the best.)
}

\resizebox{\textwidth}{!}{%
\begin{tabular}{
@{}l
ccc @{\hspace{10pt}}
ccc @{\hspace{10pt}}
ccc@{}
}
\toprule
\multirow{2}{*}{\textbf{Conditions}} &
\multicolumn{3}{c}{\textbf{Step1X-3D}} &
\multicolumn{3}{c}{\textbf{Hunyuan3D 2.0}} &
\multicolumn{3}{c}{\textbf{Hunyuan3D 2mini}} \\
\cmidrule(lr){2-4}\cmidrule(lr){5-7}\cmidrule(lr){8-10}

& {\tiny Uni3D-I $\uparrow$}
& {\tiny OpenShape-I $\uparrow$}
& {\tiny Size (GB) $\downarrow$}
& {\tiny Uni3D-I $\uparrow$}
& {\tiny OpenShape-I $\uparrow$}
& {\tiny Size (GB) $\downarrow$}
& {\tiny Uni3D-I $\uparrow$}
& {\tiny OpenShape-I $\uparrow$}
& {\tiny Size (GB) $\downarrow$} \\
\midrule

Original
& \colorbox{yellow}{\underline{0.3586}}
& \colorbox{yellow}{\underline{0.1480}}
& 2.452
& \colorbox{yellow}{\underline{0.3582}}
& \colorbox{yellow}{0.1487}
& 2.704
& \colorbox{yellow}{\textbf{0.3614}}
& \colorbox{yellow}{\textbf{0.1490}}
& 1.042 \\

\plusmark~Pruning (random)
& 0.0829 & 0.0375 & 1.123
& 0.1171 & 0.0606 & 1.575
& 0.3084 & 0.1356 & 0.954 \\

\plusmark~Vitality-Aware
& \colorbox{yellow}{0.3584} & 0.1472 & 1.123
& \colorbox{yellow}{0.3576}
& \colorbox{yellow}{\textbf{0.1491}} & 1.575
& 0.3437 & 0.1417 & 0.954 \\

\plusmark~Quantization (4b)
& 0.3489 & 0.1466 & \textbf{0.803}
& 0.3134 & 0.1351 & \textbf{0.709}
& 0.3356 & 0.1399 & \textbf{0.442} \\

\;\;\;\;\,Quantization (8b)
& \colorbox{yellow}{\textbf{0.3601}}
& \colorbox{yellow}{0.1479}
& 0.910
& \colorbox{yellow}{0.3574}
& \colorbox{yellow}{\underline{0.1488}}
& 1.031
& 0.3426 & 0.1420 & 0.622 \\

\plusmark~Adaptive Quant.
& \colorbox{yellow}{0.3579}
& \colorbox{yellow}{0.1478}
& \underline{0.843}
& 0.3528
& \colorbox{yellow}{0.1480}
& \underline{0.909}
& 0.3437 & 0.1425 & \underline{0.578} \\

\midrule
\plusmark~Fine-tuning (\textbf{Ours})
& \colorbox{yellow}{0.3580}
& \colorbox{yellow}{\textbf{0.1489}}
& \makecell[c]{\underline{0.843}\\{\tiny($-$\textbf{65.63\%})}}
& \colorbox{yellow}{\textbf{0.3601}}
& \colorbox{yellow}{\textbf{0.1491}}
& \makecell[c]{\underline{0.909}\\{\tiny($-$\textbf{66.37\%})}}
& \colorbox{yellow}{\underline{0.3608}}
& \colorbox{yellow}{\underline{0.1484}}
& \makecell[c]{\underline{0.578}\\{\tiny($-$\textbf{44.50\%})}} \\

\bottomrule
\end{tabular}
}

\label{tab:ablation}
\end{table*}

\subsubsection{Quantitative Ablation Study}
For detailed evaluation of our proposed components, we show quantitative measurement in \cref{tab:ablation}.
To evaluate the versatility of our proposed method, we conduct ablation study on 3 different models of Step1X-3D, Hunyuan3D 2.0, and Hunyuan3D 2mini.
Starting from the full-parameter original model, we first show the output from random layer pruned model (\plusmark~Pruning (random)).
Since many vital layers are removed, the overall quality of model is significantly degraded. Then we apply our vitality-aware pruning strategy, where we prune only non-vital layers (\plusmark~Vitality-Aware).
With removing the redundant layers, we can dramatically remove the model size with minimal performance drop.
This result clearly show the effectiveness of our proposed pruning stage.

With layer pruned model, we apply quantization to remaining layers (\plusmark~Quantization). With 8-bit quantization, we can further reduce the model size, and the performance is slightly degraded or similar to the original model.
However, with 4-bit quantization, we can see the model size is further decreased but the quality of the model has been dropped, especially for the Hunyuan3D models.
With applying our proposed adaptive quantization(\plusmark~Adaptive Quant), we can further reduce the model from 8-bit quantization while minimizing the performance drop.
After using our fine-tuning strategy (\plusmark~Fine-tuning), we are able to achieve performance of the compressed model that was nearly identical to that of the full-parameter model.
In the case of Step1X-3D, the difference between the vital and non-vital layers is clear, therefore we can obtain a good model during the pruning step and fine-tuning had little effect.

\begin{figure*}[t!]
    \centering
    \includegraphics[trim={3mm 0mm 0.5mm 2mm}, clip, width=\linewidth]{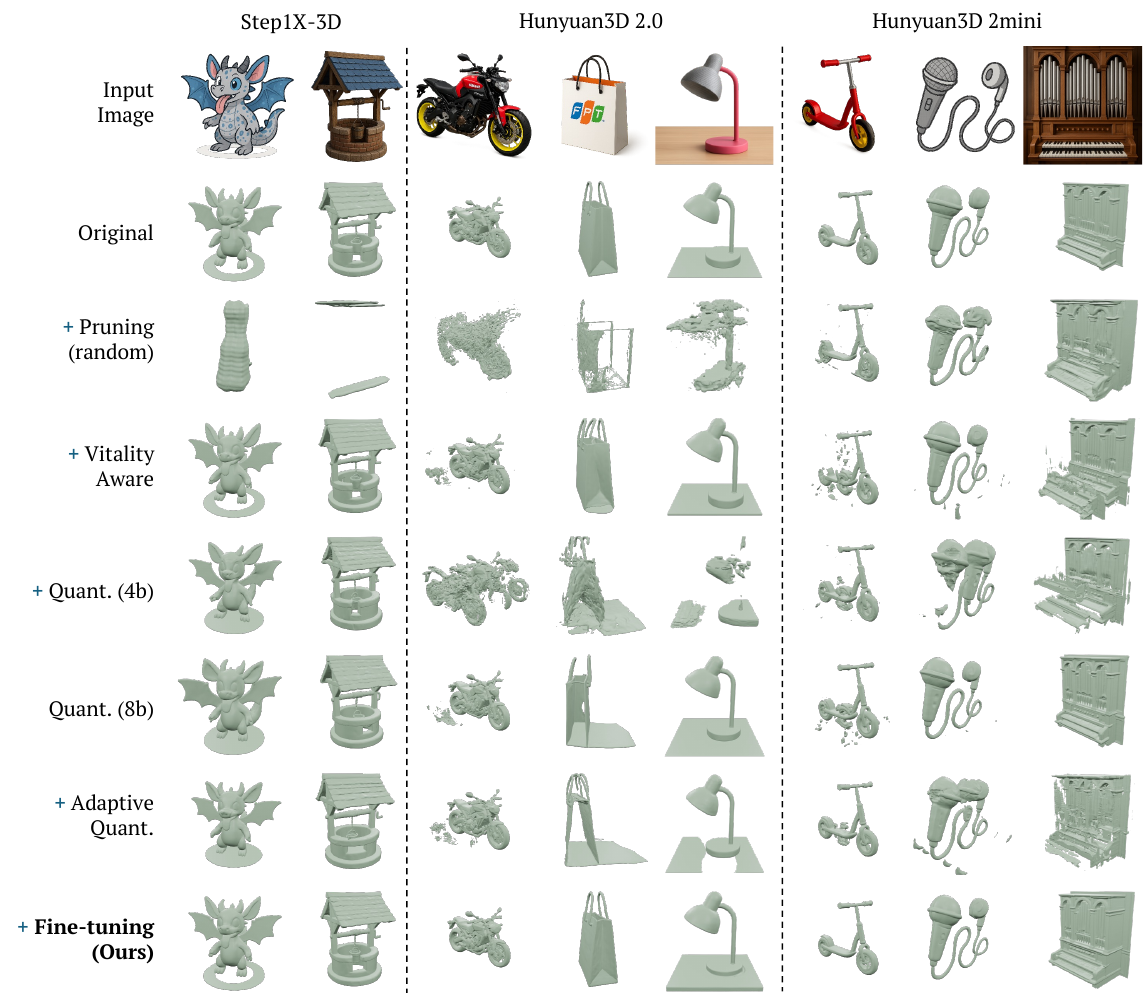}
    \caption{\textbf{Qualitative Comparisons on Ablation Study.} 
    Random pruning severely degrades performance, whereas pruning non-vital layers incurs only minor quality loss.
    Adaptive quantization preserves near 8-bit quality despite a smaller model size, and fine-tuning further restores performance to nearly the original level.
    }
    \label{fig:ablation_qualitative}
\end{figure*}

\subsubsection{Qualitative Ablation Study}
To clearly demonstrate the effect of each step in our method, we provide qualitative comparisons as shown in \cref{fig:ablation_qualitative}.
The model with only random pruning applied shows severe degradation.
With vitality-aware pruning, performance remains similar to the original, though artifacts appear in the Hunyuan3D models.
Under uniform 4-bit quantization, performance drops while quality is partially restored when applying our adaptive quantization.
Nevertheless, the Hunyuan3D models still exhibit artifacts. After fine-tuning, all models achieve results almost identical to those of the full-parameter models.

\section{Exploring Applicability}
\subsection{Experiment on TRELLIS}
\label{sec:trellis_applicability}
To further demonstrate the generality of our approach, we apply our compression method to TRELLIS~\cite{xiang2024structured}, a DiT-based 3D generative model with a different pipeline from Step1X-3D and Hunyuan3D 2.0.
TRELLIS consists of two flow stages: the Sparse Structure Flow (SSF), which primarily governs geometry generation, and the SLAT Flow, which mainly affects texture and appearance.
We compress only the SSF, leaving the appearance-oriented SLAT Flow as future work.
A detailed vitality analysis of both flows is provided in the appendix.

\begin{figure*}[t]
    \centering
    \includegraphics[trim={0mm 7mm 0mm 11mm}, clip, width=0.85\linewidth]{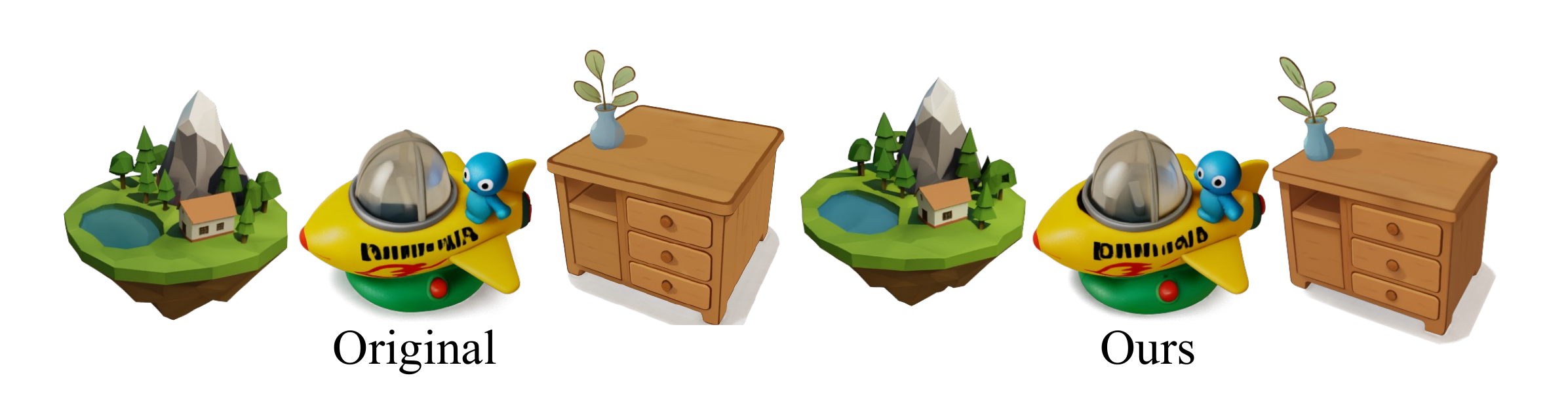}
    \caption{{%
        \textbf{Qualitative Results of Applying Our Method to TRELLIS.} Our approach can also effectively compress the model while preserving the original synthesis quality.
    }}
    \label{fig:main_trellis}
\end{figure*}

Using the same vitality-guided pruning and adaptive quantization strategy as in our main experiments, we compress the SSF with $\tau = 0.136$ and $\tau_{\text{quant}} = 0.148$, pruning 6 of 24 layers, quantizing 9 low-vitality layers to 4-bit precision, and the remaining layers to 8-bit precision.
As shown in \cref{fig:main_trellis} and \cref{tab:main_trellis}, the compressed SSF reduces the model size to 29.7\% of the original with only a 1.6\% performance drop and negligible visual differences, demonstrating that our method generalizes well to diverse DiT-based 3D generative models.

\begin{table*}[t]
\centering
\setlength{\tabcolsep}{3pt}
\renewcommand{\arraystretch}{1.05}
\scriptsize
\caption{
    \textbf{TRELLIS~(SSF) Compression Quantitative Results.}
    Applying our vitality-guided compression to the SSF of TRELLIS substantially reduces model size, latency, and peak VRAM while maintaining comparable generation quality.
}
\begin{tabularx}{\columnwidth}{
    >{\RaggedRight\arraybackslash}m{0.1\columnwidth}
    >{\centering\arraybackslash}m{0.16\columnwidth}
    >{\centering\arraybackslash}m{0.16\columnwidth}
    >{\centering\arraybackslash}m{0.16\columnwidth}
    >{\centering\arraybackslash}m{0.16\columnwidth}
    >{\centering\arraybackslash}m{0.16\columnwidth}
}
\toprule
\textbf{Method} &
\textbf{Uni3D-I} &
\textbf{Model Size (GB)} &
\textbf{Per-Step TFLOPs} &
\textbf{Latency (s)} &
\textbf{Peak VRAM (GB)} \\
\midrule
Original                                    & 0.3442 & 1.078 & 5.23 & 3.716 & 1.188 \\
\plusmark~Ours                              & 0.3395 \tiny{\textbf{($-$1.6\%)}}& 0.320 \tiny{\textbf{($-$70.3\%)}} & 3.92 \tiny{\textbf{($\times$1.33)}} & 2.802 \tiny{\textbf{($\times$1.33)}} & 0.442 \tiny{\textbf{($-$62.8\%)}} \\
\bottomrule
\end{tabularx}
\label{tab:main_trellis}
\end{table*}

\begin{figure*}[t]
    \centering
    \includegraphics[width=\linewidth]{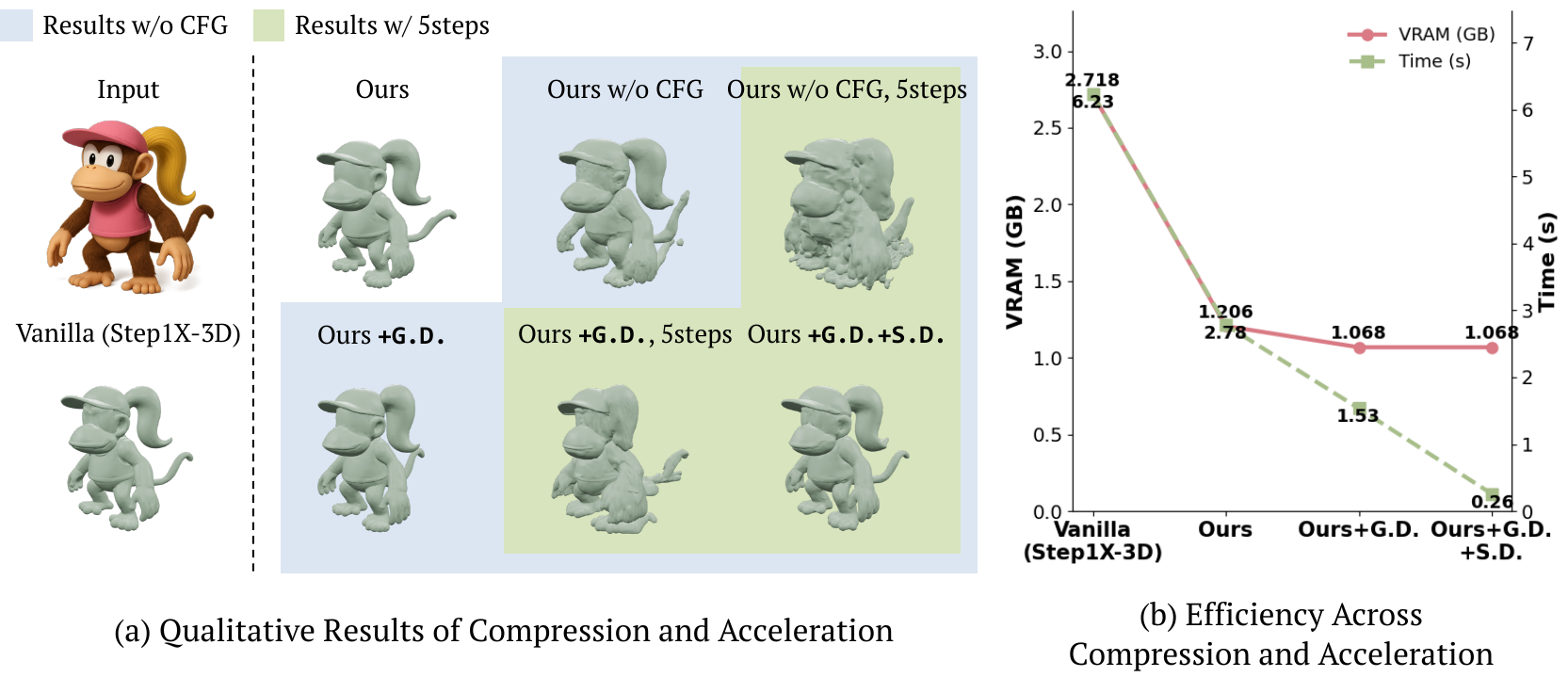}
    \caption{\textbf{Results of Applying Additional Distillation for Acceleration}.
    \textbf{(a)} We further accelerate the pipeline by applying additional distillation strategies to remove guidance and reduce the number of inference steps in the denoising process, similar to FlashVDM~\cite{lai2025unleashing}.
    \textbf{(b)} Our compression approach mainly reduces memory usage, while guidance and step distillation progressively decrease runtime without sacrificing stability.
    }
    \label{fig:ablation_acc}
\end{figure*}

\subsection{Secondary Distillation for Acceleration}
Our approach is complementary to inference-time distillation, enabling both lightweight and fast 3D shape generation.
To further accelerate denoising, we adopt progressive distillation consisting of guidance distillation to remove classifier-free guidance (CFG) and step distillation to reduce the number of denoising steps. Implementation details are provided in the appendix.

\Cref{fig:ablation_acc}(a) presents qualitative comparisons of the original model, our compressed model (\texttt{Ours}), and progressively accelerated variants with guidance distillation (\texttt{G.D.}) and step distillation (\texttt{S.D.}).
Despite removing CFG and reducing denoising to only five steps, the proposed acceleration pipeline preserves stable generation quality.

As shown in \Cref{fig:ablation_acc}(b), structural compression reduces peak memory usage by 2.5$\times$, while guidance distillation and step distillation further achieve 2.6$\times$ and up to 13.5$\times$ inference speedups, respectively, without additional memory overhead.
These results demonstrate that structural compression and progressive distillation are highly complementary, enabling efficient and scalable DiT-based image-to-3D generation.
Additional comparisons are provided in the appendix.

\section{Conclusion}
In this work, we address the challenge of reducing the computational burden of large image-to-shape DiTs while maintaining high synthesis quality.
We present a vitality-aware compression framework that integrates layer pruning, adaptive quantization, and targeted fine-tuning to systematically reduce model complexity.
Through extensive experiments on state-of-the-art architectures, including Step1X-3D, Hunyuan3D 2.0, and Hunyuan3D 2mini, our approach achieves over 50\% reduction in model size with minimal degradation in 3D shape fidelity.
These results highlight that analyzing layer vitality effectively identifies structural redundancies within DiT architectures, enabling substantial compression while avoiding performance degradation in 3D shape synthesis.
Our approach, as the first approach for physical model compression of DiTs for 3D geometry synthesis, opens up new possibilities for scalable, plug-and-play 3D generation in resource-constrained and interactive environments.

\newpage
\clearpage

\bibliographystyle{splncs04}
\bibliography{main}

\begin{thebibliography}{10}
\providecommand{\url}[1]{\texttt{#1}}
\providecommand{\urlprefix}{URL }
\providecommand{\doi}[1]{https://doi.org/#1}

\bibitem{avrahami2025stable}
Avrahami, O., Patashnik, O., Fried, O., Nemchinov, E., Aberman, K., Lischinski, D., Cohen-Or, D.: {Stable Flow}: Vital layers for training-free image editing. In: CVPR (2025)

\bibitem{betker2023improving}
Betker, J., Goh, G., Jing, L., Brooks, T., Wang, J., Li, L., Ouyang, L., Zhuang, J., Lee, J., Guo, Y., et~al.: Improving image generation with better captions. Computer Science. https://cdn. openai. com/papers/dall-e-3. pdf  \textbf{2}(3), ~8 (2023)

\bibitem{caron2021emerging}
Caron, M., Touvron, H., Misra, I., J\'egou, H., Mairal, J., Bojanowski, P., Joulin, A.: Emerging properties in self-supervised vision transformers. In: ICCV (2021)

\bibitem{chan2022efficient}
Chan, E.R., Lin, C.Z., Chan, M.A., Nagano, K., Pan, B., De~Mello, S., Gallo, O., Guibas, L.J., Tremblay, J., Khamis, S., et~al.: Efficient geometry-aware 3d generative adversarial networks. In: Proceedings of the IEEE/CVF conference on computer vision and pattern recognition. pp. 16123--16133 (2022)

\bibitem{chan2021pi}
Chan, E.R., Monteiro, M., Kellnhofer, P., Wu, J., Wetzstein, G.: pi-gan: Periodic implicit generative adversarial networks for 3d-aware image synthesis. In: Proceedings of the IEEE/CVF conference on computer vision and pattern recognition. pp. 5799--5809 (2021)

\bibitem{chang2024sparsedit}
Chang, S., Wang, P., Tang, J., Wang, F., Yang, Y.: Sparsedit: Token sparsification for efficient diffusion transformer. arXiv preprint arXiv:2412.06028  (2024)

\bibitem{chen2025q}
Chen, L., Meng, Y., Tang, C., Ma, X., Jiang, J., Wang, X., Wang, Z., Zhu, W.: {Q-DiT}: Accurate post-training quantization for diffusion transformers. In: CVPR (2025)

\bibitem{chen2025primx}
Chen, Z., Tang, J., Dong, Y., Cao, Z., Hong, F., Lan, Y., Wang, T., Xie, H., Wu, T., Saito, S., Pan, L., Lin, D., Liu, Z.: 3dtopia-xl: High-quality 3d pbr asset generation via primitive diffusion. In: CVPR (2025)

\bibitem{chou2023diffusionsdf}
Chou, G., Bahat, Y., Heide, F.: {Diffusion-SDF}: Conditional generative modeling of signed distance functions. In: ICCV (2023)

\bibitem{deitke2023objaverse}
Deitke, M., Schwenk, D., Salvador, J., Weihs, L., Michel, O., VanderBilt, E., Schmidt, L., Ehsani, K., Kembhavi, A., Farhadi, A.: Objaverse: A universe of annotated {3D} objects. In: CVPR (2023)

\bibitem{fan2019reducing}
Fan, A., Grave, E., Joulin, A.: Reducing transformer depth on demand with structured dropout. In: ICLR (2020)

\bibitem{fang2025tinyfusion}
Fang, G., Li, K., Ma, X., Wang, X.: {TinyFusion}: Diffusion transformers learned shallow. In: CVPR (2025)

\bibitem{fang2023structural}
Fang, G., Ma, X., Wang, X.: Structural pruning for diffusion models. In: Advances in Neural Information Processing Systems (2023)

\bibitem{henzler2019platonicgan}
Henzler, P., Mitra, N.J., Ritschel, T.: {Escaping Plato's Cave}: {3D} shape from adversarial rendering. In: The IEEE International Conference on Computer Vision (ICCV) (2019)

\bibitem{hong2023debiasing}
Hong, S., Ahn, D., Kim, S.: Debiasing scores and prompts of 2d diffusion for view-consistent text-to-3d generation. Advances in Neural Information Processing Systems  \textbf{36},  11970--11987 (2023)

\bibitem{hong2023lrm}
Hong, Y., Zhang, K., Gu, J., Bi, S., Zhou, Y., Liu, D., Liu, F., Sunkavalli, K., Bui, T., Tan, H.: Lrm: Large reconstruction model for single image to 3d. arXiv preprint arXiv:2311.04400  (2023)

\bibitem{hong2024lrm}
Hong, Y., Zhang, K., Gu, J., Bi, S., Zhou, Y., Liu, D., Liu, F., Sunkavalli, K., Bui, T., Tan, H.: {LRM}: Large reconstruction model for single image to {3D}. In: ICLR (2024)

\bibitem{hu2025turbo3d}
Hu, H., Yin, T., Luan, F., Hu, Y., Tan, H., Xu, Z., Bi, S., Tulsiani, S., Zhang, K.: {Turbo3D}: Ultra-fast text-to-{3D} generation. In: CVPR (2025)

\bibitem{hui2022neuralwavelet}
Hui, K.H., Li, R., Hu, J., Fu, C.W.: Neural wavelet-domain diffusion for {3D} shape generation (2022)

\bibitem{hwang2025tq}
Hwang, Y., Lee, H., Kang, J.: {TQ-DiT}: Efficient time-aware quantization for diffusion transformers. arXiv preprint arXiv:2502.04056  (2025)

\bibitem{jiao2019tinybert}
Jiao, X., Yin, Y., Shang, L., Jiang, X., Chen, X., Li, L., Wang, F., Liu, Q.: {TinyBERT}: Distilling bert for natural language understanding. arXiv preprint arXiv:1909.10351  (2019)

\bibitem{jun2023shap}
Jun, H., Nichol, A.: Shap-e: Generating conditional 3d implicit functions. arXiv preprint arXiv:2305.02463  (2023)

\bibitem{kim2025tv}
Kim, M.J., Kim, D., Yun, S., Choo, J.: {TV-LiVE}: Training-free, text-guided video editing via layer informed vitality exploitation. arXiv preprint arXiv:2506.07205  (2025)

\bibitem{lai2025unleashing}
Lai, Z., Zhao, Y., Zhao, Z., Liu, H., Wang, F., Shi, H., Yang, X., Lin, Q., Huang, J., Liu, Y., et~al.: Unleashing vecset diffusion model for fast shape generation. In: ICCV (2025)

\bibitem{lan2024ga}
Lan, Y., Zhou, S., Lyu, Z., Hong, F., Yang, S., Dai, B., Pan, X., Loy, C.C.: Gaussiananything: Interactive point cloud latent diffusion for 3d generation. In: ICLR (2025)

\bibitem{lee2024dit}
Lee, Y., Lee, Y.J., Hwang, S.J.: {Dit-Pruner}: Pruning diffusion transformer models for text-to-image synthesis using human preference scores. In: ECCV (2024)

\bibitem{li2024craftsman}
Li, W., Liu, J., Yan, H., Chen, R., Liang, Y., Chen, X., Tan, P., Long, X.: {CraftsMan3D}: High-fidelity mesh generation with {3D} native generation and interactive geometry refiner. In: ICLR (2024)

\bibitem{li2025step1x}
Li, W., Zhang, X., Sun, Z., Qi, D., Li, H., Cheng, W., Cai, W., Wu, S., Liu, J., Wang, Z., et~al.: {Step1X-3D}: Towards high-fidelity and controllable generation of textured {3D} assets. arXiv preprint arXiv:2505.07747  (2025)

\bibitem{liu2023openshape}
Liu, M., Shi, R., Kuang, K., Zhu, Y., Li, X., Han, S., Cai, H., Porikli, F., Su, H.: {OpenShape}: Scaling up 3d shape representation towards open-world understanding. In: NeurIPS (2023)

\bibitem{liu2023one}
Liu, M., Xu, C., Jin, H., Chen, L., Varma~T, M., Xu, Z., Su, H.: {One-2-3-45}: Any single image to {3D} mesh in 45 seconds without per-shape optimization. In: NeurIPS (2023)

\bibitem{luo2021dpm}
Luo, S., Hu, W.: Diffusion probabilistic models for {3D} point cloud generation. In: CVPR (2021)

\bibitem{mittal2022autosdf}
Mittal, P., Cheng, Y.C., Singh, M., Tulsiani, S.: {AutoSDF}: Shape priors for {3D} completion, reconstruction and generation. In: CVPR (2022)

\bibitem{nash2020polygen}
Nash, C., Ganin, Y., Eslami, S.M.A., Battaglia, P.W.: {PolyGen}: An autoregressive generative model of {3D} meshes. In: ICML (2020)

\bibitem{peruzzo2025adaptor}
Peruzzo, E., Karjauv, A., Sebe, N., Ghodrati, A., Habibian, A.: Adaptor: Adaptive token reduction for video diffusion transformers. In: Proceedings of the IEEE/CVF Conference on Computer Vision and Pattern Recognition (CVPR) Workshops (2025)

\bibitem{sanh2019distilbert}
Sanh, V., Debut, L., Chaumond, J., Wolf, T.: {DistilBERT, a distilled version of BERT}: smaller, faster, cheaper and lighter. arXiv preprint arXiv:1910.01108  (2019)

\bibitem{shen2020q}
Shen, S., Dong, Z., Ye, J., Ma, L., Yao, Z., Gholami, A., Mahoney, M.W., Keutzer, K.: Q-bert: Hessian based ultra low precision quantization of bert. In: AAAI (2020)

\bibitem{shue2023triplanediffusion}
Shue, J.R., Chan, E.R., Po, R., Ankner, Z., Wu, J., Wetzstein, G.: {3D} neural field generation using triplane diffusion. In: CVPR (2023)

\bibitem{siddiqui_meshgpt_2024}
Siddiqui, Y., Alliegro, A., Artemov, A., Tommasi, T., Sirigatti, D., Rosov, V., Dai, A., Nie{\ss}ner, M.: {MeshGPT}: Generating triangle meshes with decoder-only transformers. In: CVPR (2024)

\bibitem{szymanowicz24splatter}
Szymanowicz, S., Rupprecht, C., Vedaldi, A.: {Splatter Image}: Ultra-fast single-view {3D} reconstruction. In: CVPR (2024)

\bibitem{tang2024lgm}
Tang, J., Chen, Z., Chen, X., Wang, T., Zeng, G., Liu, Z.: {LGM}: Large multi-view gaussian model for high-resolution {3D} content creation. In: ECCV (2024)

\bibitem{TripoSR2024}
Tochilkin, D., Pankratz, D., Liu, Z., Huang, Z., , Letts, A., Li, Y., Liang, D., Laforte, C., Jampani, V., Cao, Y.P.: {TripoSR}: Fast {3D} object reconstruction from a single image. arXiv preprint arXiv:2403.02151  (2024)

\bibitem{vahdat2022lion}
Vahdat, A., Williams, F., Gojcic, Z., Litany, O., Fidler, S., Kreis, K., et~al.: {LION}: Latent point diffusion models for {3D} shape generation. In: NeurIPS (2022)

\bibitem{wang2020minilm}
Wang, W., Wei, F., Dong, L., Bao, H., Yang, N., Zhou, M.: {MiniLM}: Deep self-attention distillation for task-agnostic compression of pre-trained transformers. In: NeurIPS (2020)

\bibitem{wu20163dgan}
Wu, J., Zhang, C., Xue, T., Freeman, B., Tenenbaum, J.: Learning a probabilistic latent space of object shapes via 3d generative-adversarial modeling. In: NeurIPS (2016)

\bibitem{wu2024ptq4dit}
Wu, J., Wang, H., Shang, Y., Shah, M., Yan, Y.: {PTQ4DiT}: Post-training quantization for diffusion transformers. In: NeurIPS (2024)

\bibitem{wu2024direct3d}
Wu, S., Lin, Y., Zhang, F., Zeng, Y., Xu, J., Torr, P., Cao, X., Yao, Y.: {Direct3D}: Scalable image-to-{3D} generation via {3D} latent diffusion transformer. In: NeurIPS (2024)

\bibitem{xiang2024structured}
Xiang, J., Lv, Z., Xu, S., Deng, Y., Wang, R., Zhang, B., Chen, D., Tong, X., Yang, J.: Structured {3D} latents for scalable and versatile {3D} generation. In: CVPR (2025)

\bibitem{xie2020genvoxelnet}
Xie, J., Zheng, Z., Gao, R., Wang, W., Zhu, S.C., Wu, Y.N.: {Generative VoxelNet}: Learning energy-based models for {3D} shape synthesis and analysis. IEEE TPAMI  (2020)

\bibitem{xu2024instantmesh}
Xu, J., Cheng, W., Gao, Y., Wang, X., Gao, S., Shan, Y.: {InstantMesh}: Efficient {3D} mesh generation from a single image with sparse-view large reconstruction models. arXiv preprint arXiv:2404.07191  (2024)

\bibitem{yin2025slow}
Yin, T., Zhang, Q., Zhang, R., Freeman, W.T., Durand, F., Shechtman, E., Huang, X.: From slow bidirectional to fast autoregressive video diffusion models. In: CVPR (2025)

\bibitem{you2025layer}
You, H., Barnes, C., Zhou, Y., Kang, Y., Du, Z., Zhou, W., Zhang, L., Nitzan, Y., Liu, X., Lin, Z., et~al.: Layer-and timestep-adaptive differentiable token compression ratios for efficient diffusion transformers. In: CVPR (2025)

\bibitem{yuan2023goae}
Yuan, Z., Zhu, Y., Li, Y., Liu, H., Yuan, C.: Make encoder great again in 3d gan inversion through geometry and occlusion-aware encoding. In: ICCV (2023)

\bibitem{zafrir2019q8bert}
Zafrir, O., Boudoukh, G., Izsak, P., Wasserblat, M.: {Q8BERT}: Quantized 8bit bert. In: 2019 Fifth Workshop on Energy Efficient Machine Learning and Cognitive Computing-NeurIPS Edition (EMC2-NIPS) (2019)

\bibitem{zhang2022meshinversion}
Zhang, J., Ren, D., Cai, Z., Yeo, C.K., Dai, B., Loy, C.C.: Monocular {3D} object reconstruction with gan inversion. In: ECCV (2022)

\bibitem{gslrm2024}
Zhang, K., Bi, S., Tan, H., Xiangli, Y., Zhao, N., Sunkavalli, K., Xu, Z.: {GS-LRM}: Large reconstruction model for {3D} gaussian splatting. In: ECCV (2024)

\bibitem{zhang2024clay}
Zhang, L., Wang, Z., Zhang, Q., Qiu, Q., Pang, A., Jiang, H., Yang, W., Xu, L., Yu, J.: {CLAY}: A controllable large-scale generative model for creating high-quality {3D} assets. ACM TOG  (2024)

\bibitem{zhao2025hunyuan3d}
Zhao, Z., Lai, Z., Lin, Q., Zhao, Y., Liu, H., Yang, S., Feng, Y., Yang, M., Zhang, S., Yang, X., et~al.: {Hunyuan3D 2.0}: Scaling diffusion models for high resolution textured {3D} assets generation. arXiv preprint arXiv:2501.12202  (2025)

\bibitem{zheng2022sdfstylegan}
Zheng, X.Y., Liu, Y., Wang, P.S., Tong, X.: {SDF-StyleGAN}: Implicit sdf-based stylegan for {3D} shape generation. In: Comput. Graph. Forum (SGP) (2022)

\bibitem{zhou2023uni3d}
Zhou, J., Wang, J., Ma, B., Liu, Y.S., Huang, T., Wang, X.: {Uni3D}: Exploring unified {3D} representation at scale. In: ICLR (2024)

\bibitem{zhou2021pvd}
Zhou, L., Du, Y., Wu, J.: {3D} shape generation and completion through point-voxel diffusion. In: CVPR (2021)

\end{thebibliography}

\newpage
\clearpage
\appendix

\makeatletter
\newcommand{\manuallabel}[2]{\def\@currentlabel{#2}\label{#1}}
\makeatother

\manuallabel{fig:user_study}{6}
\manuallabel{subsec:03-2_compression}{3.2}
\manuallabel{tab:ablation}{2}
\manuallabel{fig:ablation_qualitative}{8}
\manuallabel{sec:trellis_applicability}{5.1}

\newcommand{\refofpaper}[1]{of the main paper}
\newcommand{\refinpaper}[1]{in the main paper}

\renewcommand{\thesection}{\Alph{section}}
\renewcommand{\thetable}{\Alph{table}}
\renewcommand{\thefigure}{\Alph{figure}}
\setcounter{section}{0}
\setcounter{table}{0}
\setcounter{figure}{0}

In this appendix, we provide additional experimental details (\cref{sec:app_exp_detail}), user study settings (\cref{sec:app_userstudy}), and supplementary methodological explanations (\cref{sec:app_methodology}). 
We further present extended results for baseline comparisons (\cref{sec:app_add_baseline_comparison}), comparisons with existing DiT compression methods (\cref{sec:app_comp_diff_dit}), and additional ablation studies (\cref{sec:app_add_ablation}). 
We also include a detailed analysis of vitality layers, including TRELLIS (\cref{sec:app_analysis}), and discuss limitations and future directions (\cref{sec:app_limitations}).

\section{Additional Experimental Details}
\label{sec:app_exp_detail}

\subsection{Vitality Analysis}
Our analysis requires one original inference pass and an additional inference-like pass for each layer to measure its contribution.
Consequently, the computational cost can be approximated as 
$t_{\text{infer}} \cdot (1 + N_{\text{layers}}) \cdot N_{\text{images}}$, 
where $t_{\text{infer}}$ denotes the inference time, and $N_{\text{layers}}$ and $N_{\text{images}}$ represent the number of DiT layers and the number of images used for analysis, respectively.
Thus, the overall cost scales with both the model depth and the inference efficiency of the original model.

\subsection {DiT Compression}
For Hunyuan3D 2.0, we set $\tau_d=0.18$ and $\tau_s=0.17$ for layer pruning in double-block and single-block DiT, respectively, and apply thresholds of 0.21 and 0.16 for adaptive quantization of double-block and single-block layers.
Meanwhile, since we observe that every double-block layer in Hunyuan3D 2mini plays a significant role in shape generation (\cref{fig:additional_ablation_hymini}), we do not apply layer pruning and set all layers to 8-bit in quantization except for layer 4.
For the single-block layers in the same model, we set $\tau_s=0.192$ to remove redundancy, and apply thresholds of 0.2 for single-block layers, respectively, to determine whether a layer should be assigned higher (8-bit) or lower (4-bit) bits during adaptive quantization.

For each model, the indices of the target layers (with indexing starting from 0) are as follows: Step1X-3D has target layers at index 3 for the double-block and 2 for the single-block. Hunyuan3D 2.0 has target layers at index 11 for the double-block and 26 for the single-block. Hunyuan3D 2mini has target layers at index 4 for the double-block and 12 for the single-block. 

Furthermore, we conduct model compression experiments under the following training settings: Step1X-3D is trained for 22 hours on 2 A100 GPUs with a batch size of 10 per GPU; Hunyuan3D 2.0 requires 50 hours on 2 A100 GPUs with a batch size of 3 per GPU; and Hunyuan3D 2mini is trained for 14 hours on a single A100 GPU with a batch size of 20.

\subsection{Secondary Distillation for Acceleration}
\paragraph{Guidance Distillation.}
For guidance distillation, we perform step-wise distillation between the original model and the target model without classifier-free guidance (CFG).
Specifically, the student model is trained to match the per-step latent predictions of the teacher model.
We minimize the mean squared error (MSE) between the generated latents at each diffusion step. Training is conducted using 10k images from Objaverse~\cite{deitke2023objaverse}, sampled from the same scenes used in the main method, with a learning rate of $1\times10^{-6}$.

\paragraph{Step Distillation.}
For step distillation, our goal is to match the 5-step flow of the target model to the 30-step flow of the original model.
We optimize the student model using an $\ell_1$ loss between the predicted flows of the student and teacher models.
During distillation, we use the Euler scheduler, while UniPC is used during inference.
We train the model on 48k perspectively rendered images randomly sampled from Objaverse~\cite{deitke2023objaverse} with a learning rate of $1\times10^{-6}$.

\section{User Study Details}
\label{sec:app_userstudy}
For each question, six different input image setups were presented, and participants were asked to assign a score from 1 (low) to 5 (high).
Each question included the mesh output of the original model under compression, along with results from other baselines described in Fig.~\ref{fig:user_study}~\refofpaper{}, which were randomly shuffled before being presented in the survey.
The evaluation questions are as follows:
\begin{itemize}[itemsep=1pt, topsep=1pt]
    \item \textbf{Geometric fidelity}: on a scale from 1 to 5, rate how reasonable the generated shape represents the overall geometry of the object in the input image.
    \item \textbf{Overall synthesis quality}: evaluate each generated 3D shape on a 1–5 scale, where 5 indicates highest synthesis quality and 1 indicates the lowest.
\end{itemize}

\section{Methodological Details}
\label{sec:app_methodology}

\subsection{Comparisons on Robustness of Vitality Metrics}
\label{sec:app_methodology_vitality}
\begin{table*}[t]
\centering
\scriptsize
\caption{
\textbf{Quantitative Comparison for Robustness of Vitality Metrics on double-block DiT layers of Step1X-3D.}
Comparison of Chamfer Distance (CD) and Earth Mover’s Distance (EMD)
across training scales (5k, 10k, 15k samples).
CD diff and EMD diff denote absolute percentage deviations from the 10k baseline.
}

\resizebox{\textwidth}{!}{%
\begin{tabular}{
@{}c|
cc @{\hspace{8pt}}
cccc @{\hspace{8pt}}
cccc@{}
}
\toprule
\multirow{2}{*}{\textbf{\tiny \#}} &
\multicolumn{2}{c}{\textbf{10k Points}} &
\multicolumn{4}{c}{\textbf{15k Points}} &
\multicolumn{4}{c}{\textbf{5k Points}} \\
\cmidrule(lr){2-3}
\cmidrule(lr){4-7}
\cmidrule(lr){8-11}

& CD & EMD
& CD & EMD & \textbf{CD diff (\%)} & \textbf{EMD diff (\%)}
& CD & EMD & \textbf{CD diff (\%)} & \textbf{EMD diff (\%)} \\
\midrule

0  & 0.1641 & 0.5116 & 0.1720 & 0.5159 & \textbf{4.82} & \textbf{0.82} & 0.1711 & 0.5078 & \textbf{4.28} & \textbf{0.75} \\
1  & 0.0628 & 0.3152 & 0.0759 & 0.3294 & \textbf{20.92} & \textbf{4.50} & 0.0790 & 0.3253 & \textbf{25.87} & \textbf{3.21} \\
2  & 0.0613 & 0.3270 & 0.0633 & 0.3414 & \textbf{3.31} & \textbf{4.38} & 0.0646 & 0.3281 & \textbf{5.46} & \textbf{0.31} \\
3  & 0.0160 & 0.2136 & 0.0134 & 0.2064 & \textbf{16.06} & \textbf{3.36} & 0.0134 & 0.2055 & \textbf{16.22} & \textbf{3.81} \\
4  & 0.0404 & 0.2970 & 0.0436 & 0.3037 & \textbf{7.88} & \textbf{2.28} & 0.0401 & 0.3015 & \textbf{0.74} & \textbf{1.53} \\
5  & 0.0138 & 0.2170 & 0.0179 & 0.2149 & \textbf{29.60} & \textbf{0.94} & 0.0159 & 0.2127 & \textbf{14.99} & \textbf{1.97} \\
6  & 0.1183 & 0.4822 & 0.1244 & 0.5047 & \textbf{5.18} & \textbf{4.66} & 0.1218 & 0.4833 & \textbf{3.01} & \textbf{0.23} \\
7  & 0.0012 & 0.1591 & 0.0007 & 0.1554 & \textbf{40.62} & \textbf{2.30} & 0.0013 & 0.1566 & \textbf{12.16} & \textbf{1.57} \\
8  & 0.0014 & 0.1597 & 0.0008 & 0.1594 & \textbf{40.22} & \textbf{0.16} & 0.0014 & 0.1530 & \textbf{2.67} & \textbf{4.17} \\
9  & 0.0010 & 0.1588 & 0.0006 & 0.1559 & \textbf{43.83} & \textbf{1.83} & 0.0012 & 0.1547 & \textbf{16.44} & \textbf{2.59} \\
10 & 0.0009 & 0.1591 & 0.0004 & 0.1556 & \textbf{51.70} & \textbf{2.23} & 0.0011 & 0.1568 & \textbf{30.77} & \textbf{1.46} \\
11 & 0.0010 & 0.1595 & 0.0005 & 0.1575 & \textbf{50.03} & \textbf{1.24} & 0.0012 & 0.1551 & \textbf{16.17} & \textbf{2.73} \\

\bottomrule
\end{tabular}
}
\label{tab:app_vitality}
\end{table*}

We validate the robustness of our vitality-aware metrics using the double-layer DiT block from the Step1X-3D model~\cite{li2025step1x}, with 210 images used for vitality analysis.
To assess stability across sampling densities, we vary the number of points extracted from the meshes (5k, 10k, and 15k) and report the resulting Chamfer Distance (CD) and Earth Mover’s Distance (EMD) in \cref{tab:app_vitality}.

We observe that deeper layers (\textit{e.g.}, layers 7–11) are more sensitive to sampling density, with CD values changing significantly as the sampling density varies.
This instability arises from CD’s dependence on nearest-neighbor correspondences, which makes it sensitive to sampling density and spatial distribution.
In contrast, EMD remains comparatively stable, with differences no greater than 5\% relative to our main results (measured with 10k points), even when using only 5k sampled points.
This indicates that EMD provides a more stable measure of geometry correspondence under varying sampling conditions.

Overall, these results demonstrate that the vitality-aware EMD metric remains robust across changes in sampling resolution, preserving consistent behavior at different point densities, whereas CD becomes increasingly unreliable when fewer samples are used.

\subsection{Analysis on Pruning Threshold Selection}
\label{sec:app_threshold_selection}

We further analyze how the EMD-based vitality metric behaves under different
pruning thresholds using Hunyuan3D 2.0 mini. Specifically, we vary the pruning
threshold $\tau$, where each threshold induces a different number of pruned
layers $K$. This allows us to examine how model size and performance change as
the pruning strength increases.

\begin{figure*}[t]
    \centering
    \includegraphics[trim={1mm 0mm 2mm 0mm}, clip, width=0.99\linewidth]{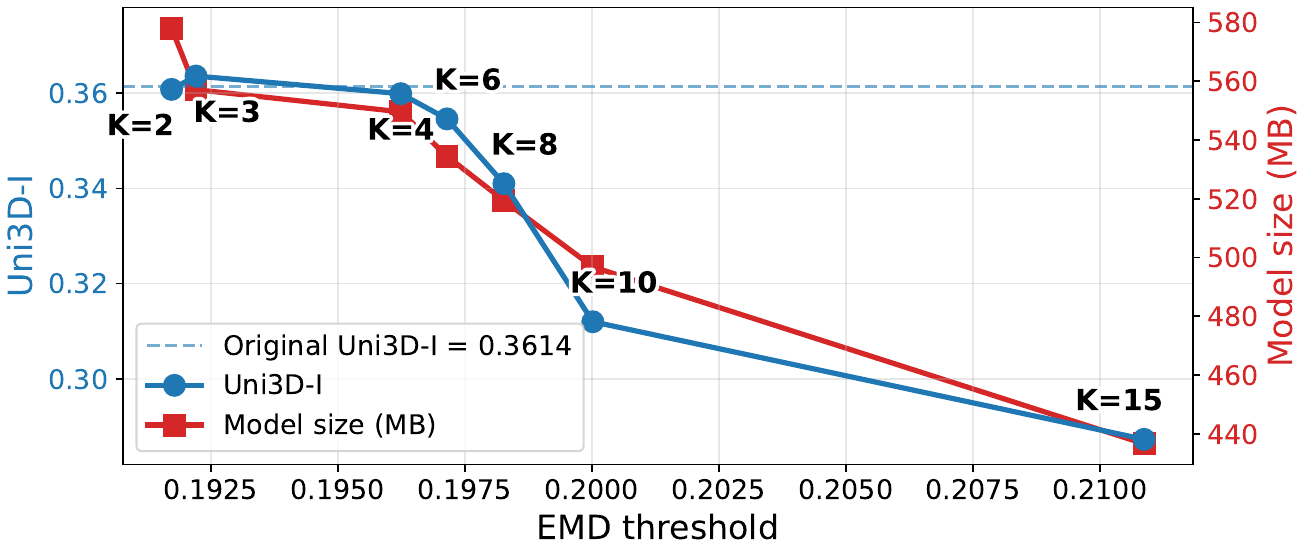}
    \caption{
        \textbf{Comparison of Different Pruning Thresholds on Hunyuan3D 2.0 mini.}
        Each threshold $\tau$ induces a different number of pruned layers $K$.
    }
    \label{fig:rebuttal_threshold}
\end{figure*}

As shown in \cref{fig:rebuttal_threshold}, increasing $\tau$ initially increases
$K$ from 2 to 6, leading to an approximately linear reduction in model size.
Despite this reduction, performance remains nearly unchanged within this range,
indicating that the layers selected by the EMD-based vitality criterion are
consistently less critical to the final output. However, when the threshold is
increased further and more than 6 layers are pruned, performance drops sharply.
This suggests that the EMD-based vitality metric provides a meaningful
separation between a safe compression regime and an over-pruning regime.

This analysis complements the sampling-density robustness study in
\cref{sec:app_methodology_vitality}. While the previous subsection shows that
EMD provides stable vitality estimates under different point sampling
resolutions, this threshold analysis shows that the resulting EMD-based pruning
criterion also produces a smooth size--performance trade-off within a moderate
pruning range and clearly exposes the point at which excessive pruning begins
to degrade performance.

\subsection{Identification of Non-Vital Layers for Pruning}
\label{sec:app_methodology_pruning}

\begin{figure}[t]
    \centering
    \includegraphics[trim={0mm 0mm 0mm 0mm}, clip, width=0.85\linewidth]{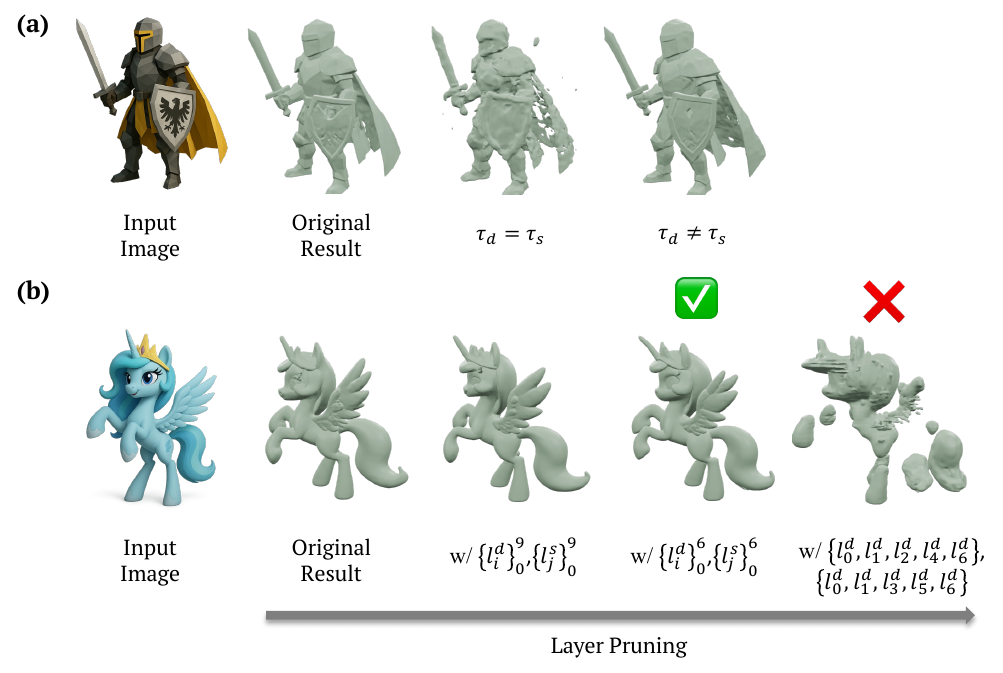}
    \caption{\textbf{Details of Layer Pruning Process.}
    \textbf{(a)} Applying identical pruning criteria to both double- and single-block layers in Hunyuan3D 2.0 causes geometric distortion.
    We therefore use distinct thresholds for the two layer types to preserve structural fidelity.
    \textbf{(b)} Layer elimination process of Step1X-3D. Minor details change below the threshold, but beyond it, the mesh structure collapses.
    Below the threshold, only fine details are altered, whereas exceeding itd causes the mesh structure to collapse.
    }
    \label{fig:app_pruning}
\end{figure}

\Cref{fig:app_pruning}~(a) shows a failure case when the same pruning criterion is applied to both double-block and single-block layers.
Specifically, we compare our method against a pruning attempt on Hunyuan3D 2.0 using a shared threshold of $\tau_{d} = \tau_{s} = 0.18$.
The geometry becomes severely distorted when applying the same standard to both layers.
Based on this observation, we adopt separate pruning criteria for double- and single-block layers.

Meanwhile, as mentioned in Sec.~\ref{subsec:03-2_compression}~\refofpaper{}, we sequentially eliminate layers beginning with those that have the lowest vitality scores, tracking how the results diverge from the baseline model output.
The procedure is illustrated in \cref{fig:app_pruning}~(b).
We observe that up to a certain threshold, only minor details are affected while the overall shape remains similar.
However, beyond this point, the mesh structure becomes completely distorted.

\subsection{Criteria for Adaptive Quantization}
\label{sec:app_methododlogy_adq}

\begin{figure}[t]
    \centering
    \includegraphics[trim={2.5mm 3mm 0mm 0mm}, clip, width=0.97\linewidth]{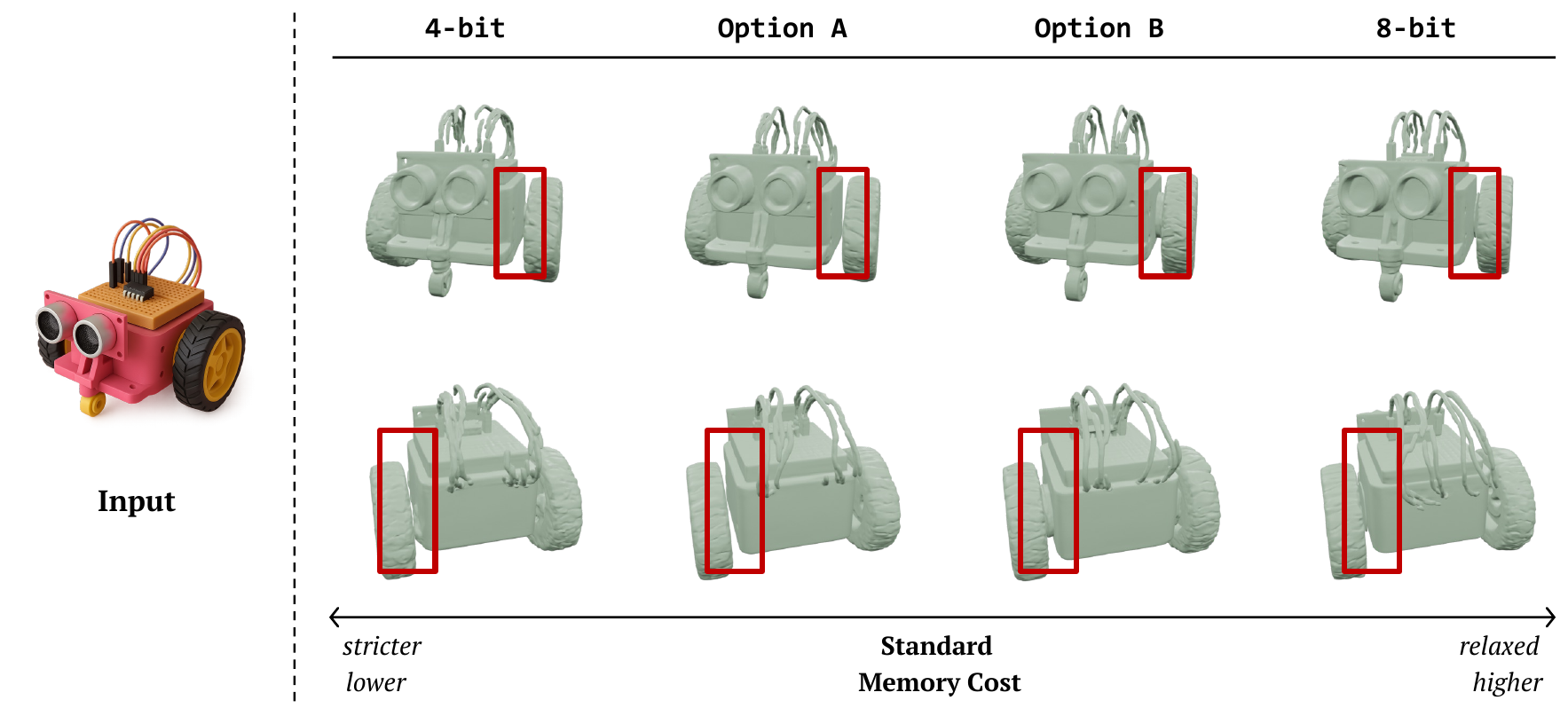}
    \caption{
    \textbf{Results of Ablation of Adaptive Quantization Strategies on Step1X-3D.}
    A stricter quantization setting in adaptive quantization leads to a more degraded initial model state.
    When comparing the marked regions across the results, a clear synthesis degradation can be observed as stricter quantization criteria are applied.
    Consequently, achieving higher compression rates at this stage requires more extensive fine-tuning under the same layer pruning configuration.
    }
    \label{fig:app_adq}
\end{figure}

We compare the results before and after fine-tuning using different adaptive quantization thresholds, as shown in \cref{fig:app_adq}.
Increasing the strictness of the threshold makes it progressively more difficult to preserve the original model performance.
Although the threshold in adaptive quantization can be freely chosen by the user, applying a stricter setting generally requires longer training or more extensive fine-tuning to maintain stability.

\begin{figure}
    \centering
    \includegraphics[trim={2mm 0mm 2mm 0mm}, clip, width=0.97\linewidth]{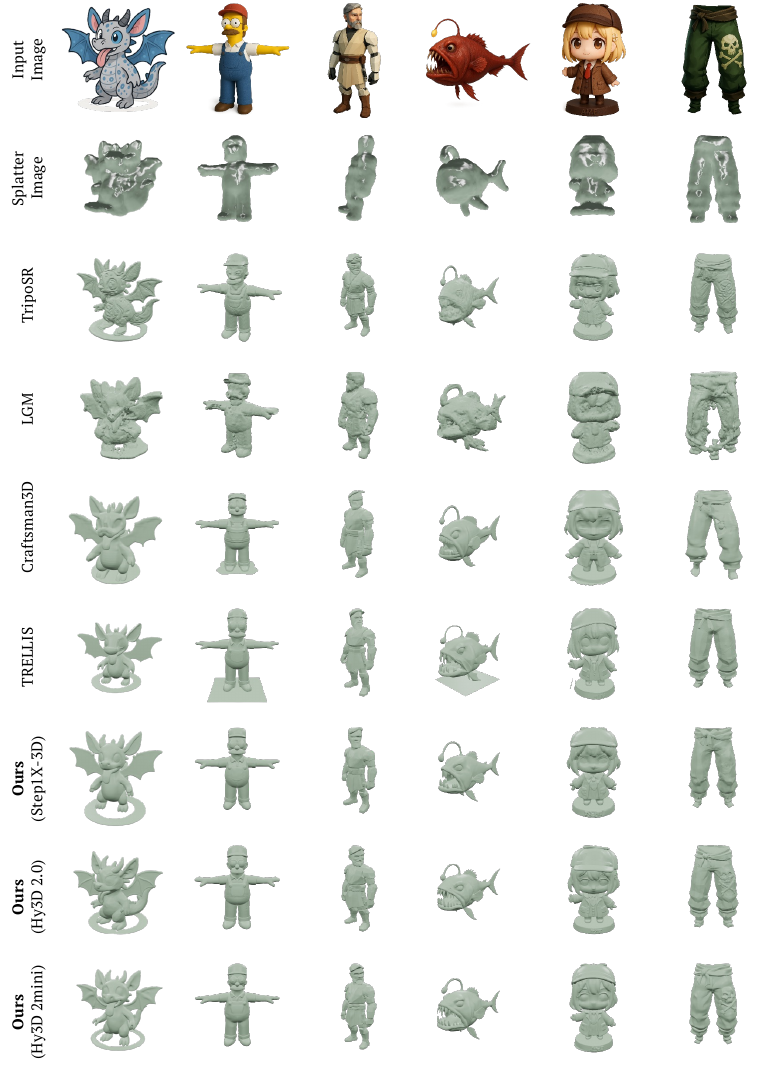}
    \caption{\textbf{Additional Qualitative Comparison with Baselines.}
    Our lightweight model generates meshes of higher quality than other baselines, similar to the original model.}
    \label{fig:additional_baseline_comparison}
\end{figure}

\section{Additional Experiment}
\label{sec:app_results}
\subsection{Qualitative Results for Baseline Comparison}
\label{sec:app_add_baseline_comparison}

Additional qualitative comparison results can be found in Fig.~\ref{fig:additional_baseline_comparison}.
This demonstrates that our approach achieves higher performance in geometry synthesis compared to existing baselines including recent DiT-based
generative models~\cite{li2024craftsman, xiang2024structured}, as the original model does.

\subsection{Qualitative Comparison with Existing DiT Compression Methods}
\label{sec:app_comp_diff_dit}

\begin{table}[t]
\centering
\scriptsize
\caption{
\textbf{Model Size Comparsion Across Difference Diffusion Model Compression Methods.}
``Size'' denotes the DiT model size, and ``Rate'' represents the compression rate after applying each method.
The percentages in Diff-Pruning represent the FFN pruning ratios based on the scoring results.
}
\resizebox{0.95\textwidth}{!}{%
\begin{tabular}{
@{}
>{\RaggedRight\arraybackslash}p{0.25\textwidth}
*{6}{>{\centering\arraybackslash}p{0.15\textwidth}}
@{}
}
\toprule
\multirow{2}{*}{\textbf{Conditions}} &
\multicolumn{2}{c}{\textbf{Step1X-3D}} &
\multicolumn{2}{c}{\textbf{Hunyuan3D 2.0}} &
\multicolumn{2}{c}{\textbf{Hunyuan3D 2mini}} \\
\cmidrule(lr){2-3}\cmidrule(lr){4-5}\cmidrule(lr){6-7}
& Size (GB) $\uparrow$ & Rate (\%) $\uparrow$
& Size (GB) $\uparrow$ & Rate (\%) $\uparrow$
& Size (GB) $\uparrow$ & Rate (\%) $\uparrow$ \\
\midrule
Vanilla & 2.452 & -- & 2.704 & -- & 1.042 & -- \\[0.15em]
TinyFusion & 1.243 & 49.31 & 1.357 & 49.82 & 0.526 & 49.52 \\[0.15em]
Diff-Pruning~(25\%) & 2.159 & 11.95 & 2.376 & 12.13 & 0.917 & 12.09 \\[0.15em]
Diff-Pruning~(60\%) & 1.748 & 28.71 & 1.922 & 28.92 & 0.741 & 28.89 \\[0.15em]
\midrule
\textbf{Ours} & 0.843 & 65.63 & 0.909 & 66.37 & 0.578 & 44.50 \\[0.15em]
\bottomrule
\end{tabular}
}
\label{tab:app_dit_compression}
\end{table}

\begin{figure}[t!]
    \centering
    \includegraphics[height=7cm, width=\linewidth]{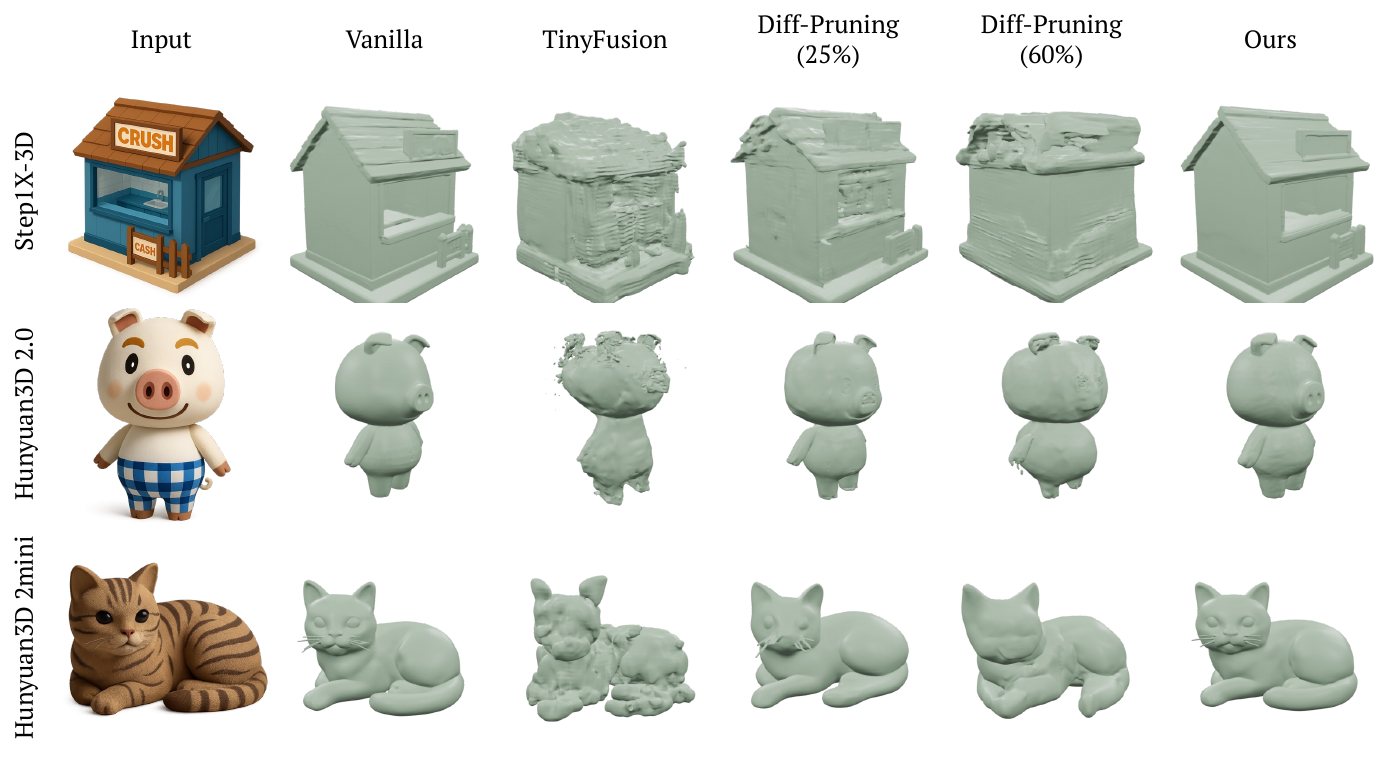}
    \caption{\textbf{Comparison with Existing Diffusion Model Compression Methods.}
    While Diff-Pruning can almost recover the performance of the original model when the FFN pruning rate is low (25\%) on Hunyuan models, unintended artifacts are observed when applying it to Step1X-3D. 
    Considering the compression rates, our approach effectively compresses the model while preserving the synthesis performance.
    }
    \label{fig:app_dit_compression}
\end{figure}

As shown in \cref{tab:app_dit_compression}, TinyFusion achieves around a 50\% compression rate, but struggles to restore generation quality across all networks, as also observed in the qualitative comparisons in \cref{fig:app_dit_compression}.
For Diff-Pruning, applying a 25\% FFN pruning ratio largely preserves performance on the Hunyuan3D models, whereas increasing the pruning ratio to 60\% leads to noticeable degradation, which is likewise reflected in \cref{fig:app_dit_compression}.
Moreover, in both settings, the overall DiT compression remains significantly smaller than that achieved by our method.

Furthermore, the observed compression results suggest that FFN layers account for only about 40--45\% of the total DiT parameters in this architecture, rather than the $\sim$66\% typically seen in standard transformers.
This is likely due to the heavier attention components in FLUX-style DiT blocks.

Overall, considering both the compression rate (\cref{tab:app_dit_compression}) and the qualitative results (\cref{fig:app_dit_compression}), our method achieves substantially higher model compression while maintaining generation quality.

\begin{table}[t]
\centering
\scriptsize
\caption{
\textbf{Additional Quantitative Comparison on Ablation Study.}
Geometric correspondence metrics (V-IoU and S-IoU) under various ablation settings
for Step1X-3D and Hunyuan3D variants. (\textbf{Bold}: best, \underline{Underline}: second best)
}
\resizebox{0.95\textwidth}{!}{%
\begin{tabular}{
@{}
>{\RaggedRight\arraybackslash}p{0.25\textwidth}
*{6}{>{\centering\arraybackslash}p{0.15\textwidth}}
@{}
}
\toprule
\multirow{2}{*}{\textbf{Conditions}} &
\multicolumn{2}{c}{\textbf{Step1X-3D}} &
\multicolumn{2}{c}{\textbf{Hunyuan3D 2.0}} &
\multicolumn{2}{c}{\textbf{Hunyuan3D 2mini}} \\
\cmidrule(lr){2-3}\cmidrule(lr){4-5}\cmidrule(lr){6-7}
& V-IoU (\%) $\uparrow$ & S-IoU (\%) $\uparrow$
& V-IoU (\%) $\uparrow$ & S-IoU (\%) $\uparrow$
& V-IoU (\%) $\uparrow$ & S-IoU (\%) $\uparrow$ \\
\midrule
Original & -- & -- & -- & -- & -- & -- \\[0.15em]
\plusmark~Pruning (random)
& 6.01 & 9.16 & 27.50 & 27.94 & 59.53 & 55.73 \\[0.15em]
\plusmark~Vitality-Aware
& \textbf{79.27} & \textbf{77.29} & \underline{71.32} & \textbf{68.66} & \textbf{74.08} & \textbf{72.05} \\[0.15em]
\plusmark~Quantization (4b)
& 62.56 & 44.69 & 51.49 & 49.49 & 69.40 & 66.34 \\[0.15em]
\;\;\;\;\,Quantization (8b)
& 69.25 & 67.09 & 69.32 & 66.64 & 73.72 & 71.71 \\[0.15em]
\plusmark~Adaptive Quant.
& 61.11 & 58.60 & 68.06 & 65.21 & 72.66 & 69.71 \\[0.15em]
\midrule
\plusmark~\textbf{Fine-tuning (Ours)}
& \underline{71.12} & \underline{68.82}
& \textbf{72.04} & \underline{68.31}
& \underline{73.77} & \underline{70.36} \\[0.15em]
\bottomrule
\end{tabular}
}
\label{tab:app_ablation}
\end{table}

\begin{table}[ht]
\centering
\setlength{\tabcolsep}{3pt}
\renewcommand{\arraystretch}{1.05}
\tiny
\caption{\textbf{Overhead of Step1X-3D Under Different Conditions.}}
\begin{tabularx}{\columnwidth}{
    >{\RaggedRight\arraybackslash}m{0.38\columnwidth}
    >{\centering\arraybackslash}m{0.24\columnwidth}
    >{\centering\arraybackslash}m{0.12\columnwidth}
    >{\centering\arraybackslash}m{0.22\columnwidth}
}
\toprule
\textbf{Conditions} &
\textbf{Per-Step TFLOPs} &
\textbf{Latency (s)} &
\textbf{Peak VRAM (GB)} \\
\midrule
Original                                    & 9.689 & 6.23 & 2.718 \\
\plusmark~Layer Puning                              & 3.774 & 2.46 & 1.417 \\
\plusmark~Quantization (4b)                         & 3.775 & 3.02 & 1.136 \\
\;\;~ Quantization (8b)              & 3.775 & 2.63 & 1.212 \\
\midrule
\textbf{\plusmark~Adaptive Quantization (Ours)}     & 3.775 & 2.78 & 1.206 \\
\bottomrule
\end{tabularx}
\label{tab:rebuttal_overhead_abl}
\end{table}

\subsection{Ablation Study}
\label{sec:app_add_ablation}

\subsubsection{Quantitative Results}

To validate geometric consistency during compression, we additionally provide quantitative ablations using volume and surface IoU metrics measured between the original and compressed models, as shown in \cref{tab:app_ablation}.
Although our compressed models achieve slightly lower performance than those using only vitality-aware layer pruning, considering the exact model size reported in Tab.~\ref{tab:ablation}~\refofpaper{} and the overall quality illustrated in Fig.~\ref{fig:ablation_qualitative}~\refofpaper{}, our method effectively restores synthesis quality while requiring minimal computational overhead. Furthermore, \cref{tab:rebuttal_overhead_abl} shows the overhead on Step1X-3D under different compression conditions.
We observe that quantization mainly improves memory efficiency, while its latency gain remains relatively modest. Since low-bit inference speed is highly implementation- and hardware-dependent, more optimized GPU-aware quantization kernels could further improve runtime performance.
All metrics are measured on an NVIDIA RTX A6000 GPU.

\subsubsection{Qualitative Results}

\begin{figure}
    \centering
    \includegraphics[trim={2mm 0mm 2mm 0mm}, clip, width=0.99\linewidth]{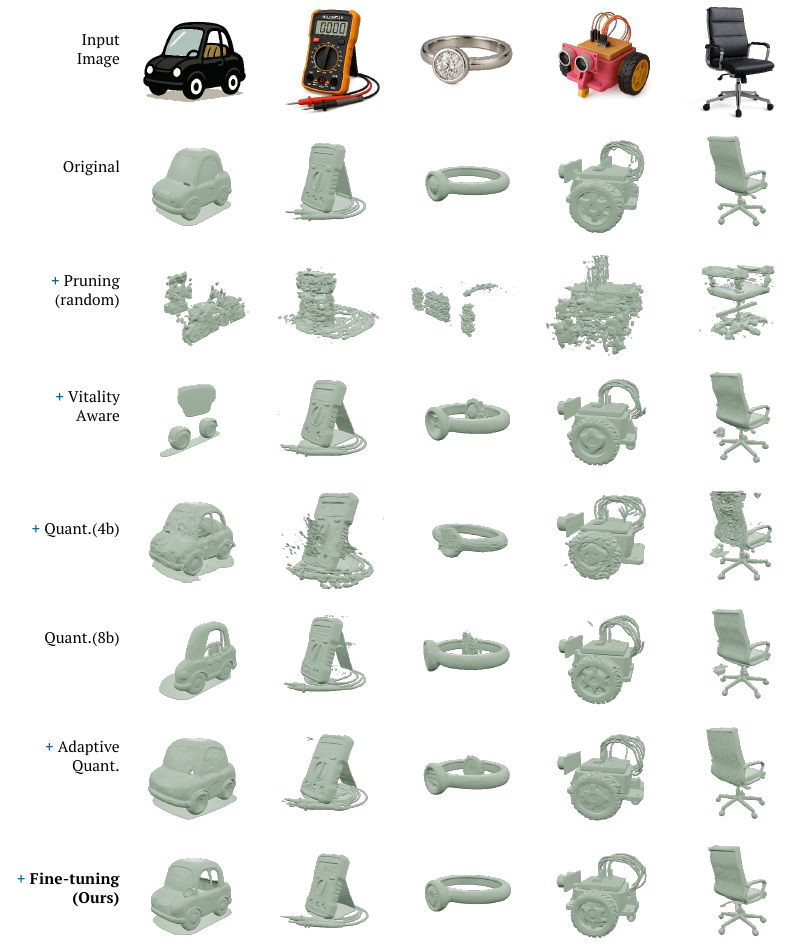}
    \caption{\textbf{Additional Qualitative Ablation Results of Hunyuan3D 2.0.} Naive pruning and quantization introduce floaters and geometry collapse, while our method preserves quality nearly identical to the original.}
    \label{fig:additional_ablation_hy3d}
\end{figure}

\begin{figure}
    \centering
    \includegraphics[trim={2.5mm 0mm 2mm 0mm}, clip, width=0.99\linewidth]{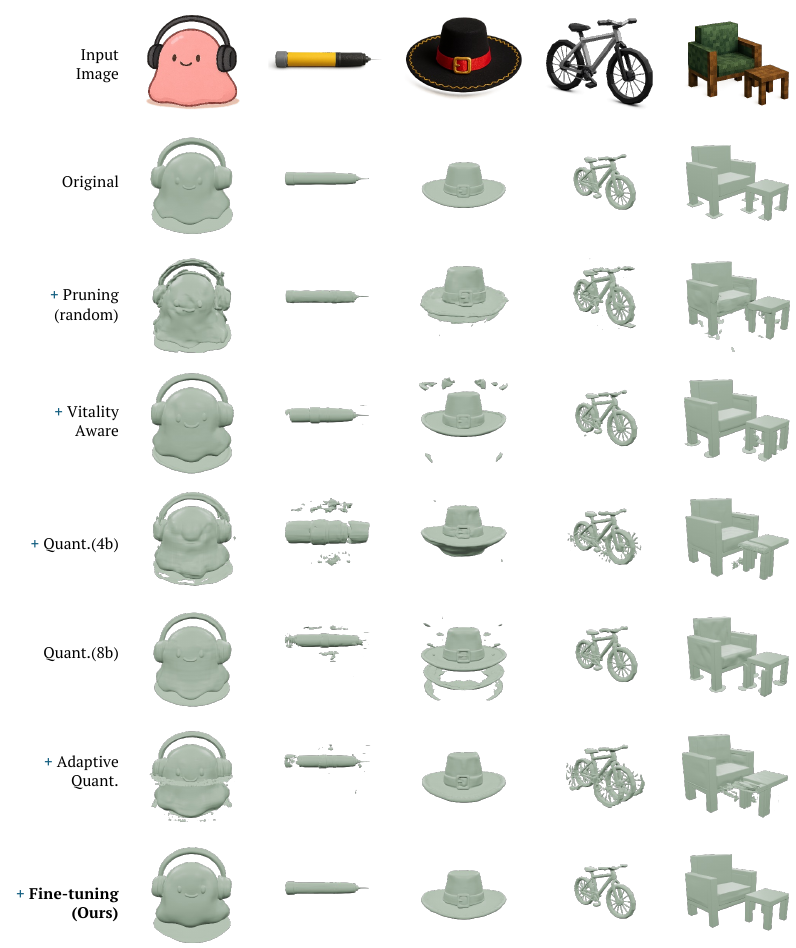}
    \caption{\textbf{Additional Qualitative Ablation Results of Hunyuan3D 2mini.} Naive pruning and quantization introduce floaters and geometry collapse, while our method preserves quality nearly identical to the original.}
    \label{fig:additional_ablation_hymini}
\end{figure}

Further qualitative ablation results for Hunyuan3D 2.0 and Hunyuan3D 2mini are presented in \cref{fig:additional_ablation_hy3d} and \cref{fig:additional_ablation_hymini}, respectively.
In Hunyuan3D models, naive pruning and quantization lead to floaters and collapsed geometry, whereas our compression method produces models that closely match the original in quality.

\subsubsection{Component-Wise Ablations Before Fine-tuning}

\begin{figure}[t]
    \centering
    \includegraphics[width=0.92\linewidth]{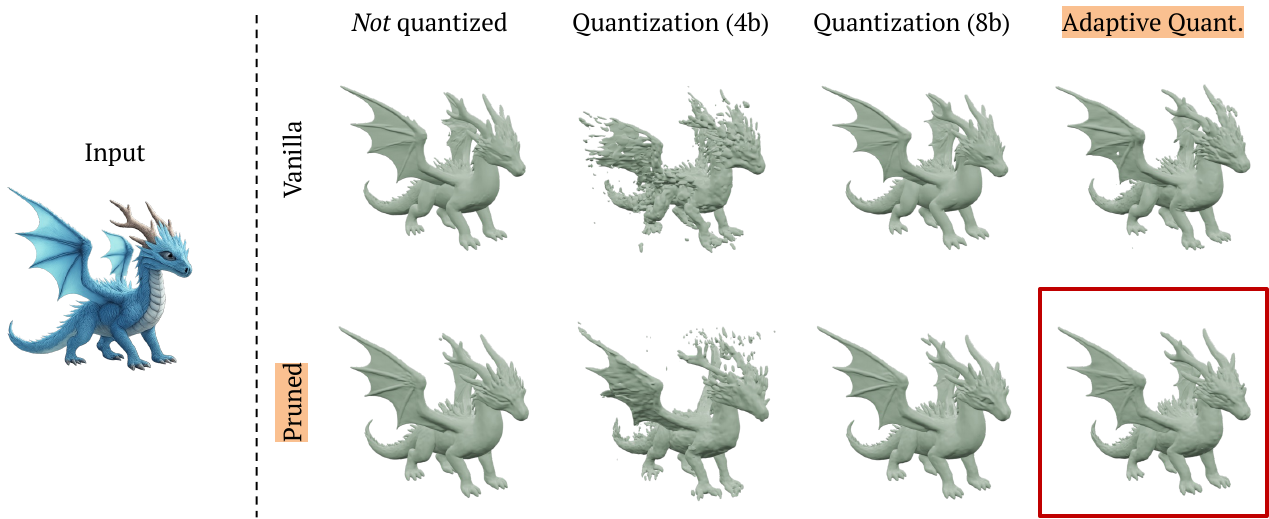}
    \caption{
    \textbf{Component-wise Ablations Before Fine-tuning on Hunyuan3D 2.0.}
    Compared with other configurations, our approach (highlighted in the red box) effectively reduces memory cost with less degradation in generation quality.
    This remaining degradation indicates the need for an additional fine-tuning stage, as used in our method.
    }
    \label{fig:app_component}
\end{figure}

\Cref{fig:app_component} visualizes the reconstruction quality after applying possible conditions of pruning and quantization on Hunyuan3D 2.0, as well as the model’s initial state before fine-tuning.
By comparing the outputs of the vanilla and pruned models across different quantization conditions, we show that our vitality-based pruning approach reduces spatial cost with minimal degradation in synthesis quality.
Furthermore, as also shown in Fig.~\ref{fig:ablation_qualitative}~\refofpaper{}, applying 4-bit quantization to all layers causes the model to struggle in forming coherent overall structures, whereas quantizing all layers to 8-bit yields output quality that is nearly identical to the non-quantized model.
In comparison to these models, our adaptive quantization strategy achieves a greater reduction in model size with substantially less degradation in performance.
Despite these improvements, a residual discrepancy remains between the outputs of the vanilla model and ours, highlighting the necessity of the fine-tuning stage.

\subsubsection{Selection of Fine-tuning Strategies}

\begin{table}[t]
\centering
\scriptsize
\caption{
\textbf{Qualitative Comparison of Ablated Fine-tuning Strategies on Hunyuan3D Models.}
Our approach yields a more stable finetuning process than alternative strategies,
leading to improved overall shape quality.
}

\resizebox{0.97\textwidth}{!}{%
\begin{tabular}{
@{}l
cccc @{\hspace{10pt}}
cccc@{}
}
\toprule
& \multicolumn{4}{c}{\textbf{Hunyuan3D 2.0}}
& \multicolumn{4}{c}{\textbf{Hunyuan3D 2mini}} \\
\cmidrule(lr){2-5}
\cmidrule(lr){6-9}

\textbf{Strategy}
& Uni3D-I $\uparrow$
& OpenShape-I $\uparrow$
& V-IoU (\%) $\uparrow$
& S-IoU (\%) $\uparrow$
& Uni3D-I $\uparrow$
& OpenShape-I $\uparrow$
& V-IoU (\%) $\uparrow$
& S-IoU (\%) $\uparrow$ \\
\midrule

Full fine-tuning
& 0.1766 & 0.0865 & 28.69 & 29.06
& 0.3210 & 0.1363 & 45.00 & 40.50 \\

w/~\textit{Max-vital}
& 0.3541 & 0.1490 & 61.50 & 56.68
& 0.3605 & 0.1479 & 66.93 & 62.28 \\

\midrule
\textbf{Ours}
& \textbf{0.3601} & \textbf{0.1491} & \textbf{72.04} & \textbf{68.31}
& \textbf{0.3608} & \textbf{0.1484} & \textbf{73.77} & \textbf{70.36} \\

\bottomrule
\end{tabular}
}
\label{tab:app_finetuning}
\end{table}

\begin{figure}[t]
    \centering
    \includegraphics[trim={2mm 0mm 0mm 0mm}, clip, width=0.9\linewidth]{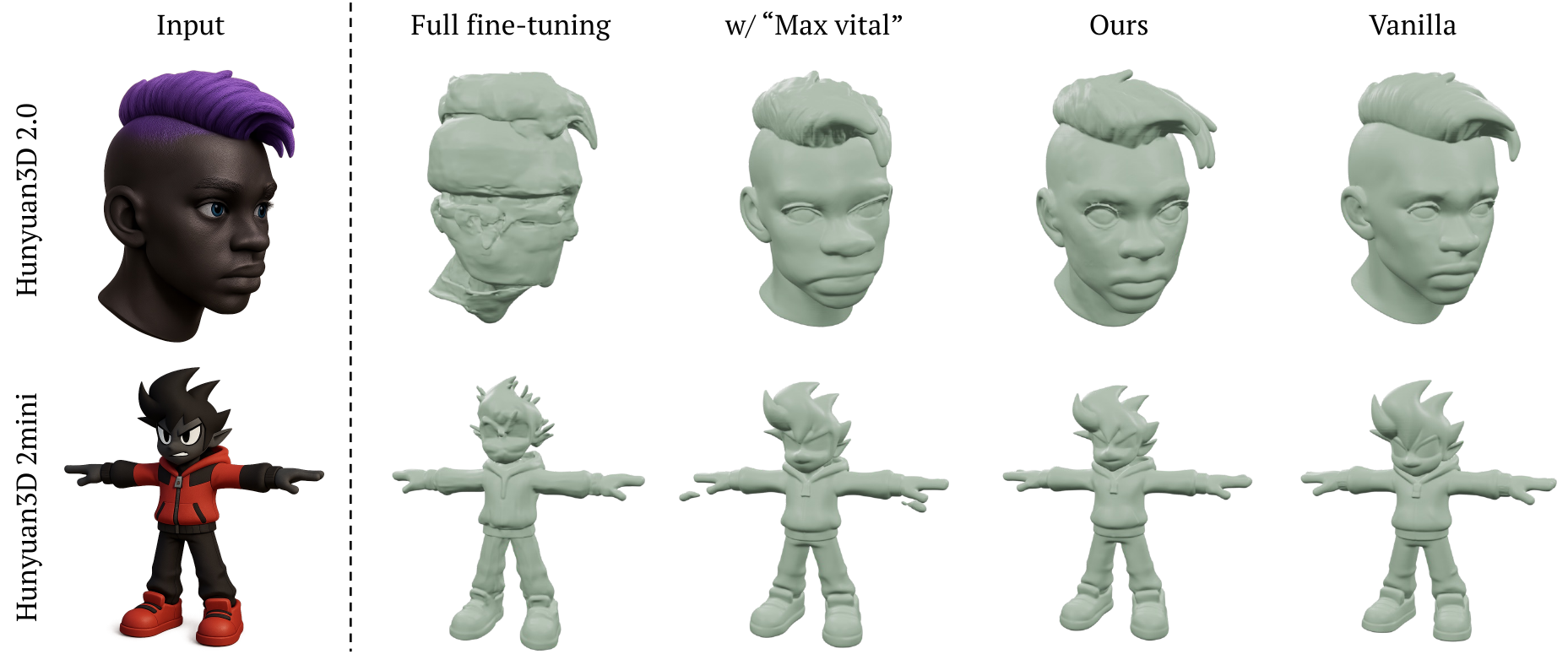}
    \caption{
    \textbf{Qualitative Ablation of Fine-tuning Strategies on Hunyuan3D models.}
    The ``Max-vital'' layers denote those with the highest vitality (\textit{i.e.}, contribute the most) per DiT block.
    We observe that fine-tuning only the lowest-vital (``Min-vital'') layers leads to more stable learning.
    }
    \label{fig:app_finetuning}
\end{figure}

To analyze the impact of different fine-tuning strategies, we conduct an ablation study on Hunyuan3D models~\cite{li2025step1x}, comparing (i) full fine-tuning, (ii) selective fine-tuning applied only to the double- and single-block layers with the highest vitality scores (\textit{i.e.}, “Max-vital” layers), and (iii) our proposed approach.
\cref{tab:app_finetuning} presents quantitative comparisons of different fine-tuning strategies on the Hunyuan3D models.
We also provide qualitative ablations of the same models in \cref{fig:app_finetuning}.
We observe that training becomes unstable when all layers of the DiT architecture are fine-tuned simultaneously.
Moreover, targeting only the “Max-vital” layers during fine-tuning often struggles to effectively mitigate degradation under compression, as it is difficult to recover finer details.
To ensure both stability and effectiveness, our approach instead focuses on the ``Min-vital'' layers.

\section{Additional Results of Secondary Distillation}
\label{sec:app_secondary_distillation}
\begin{figure}[h]
    \centering
    \includegraphics[width=0.9\linewidth]{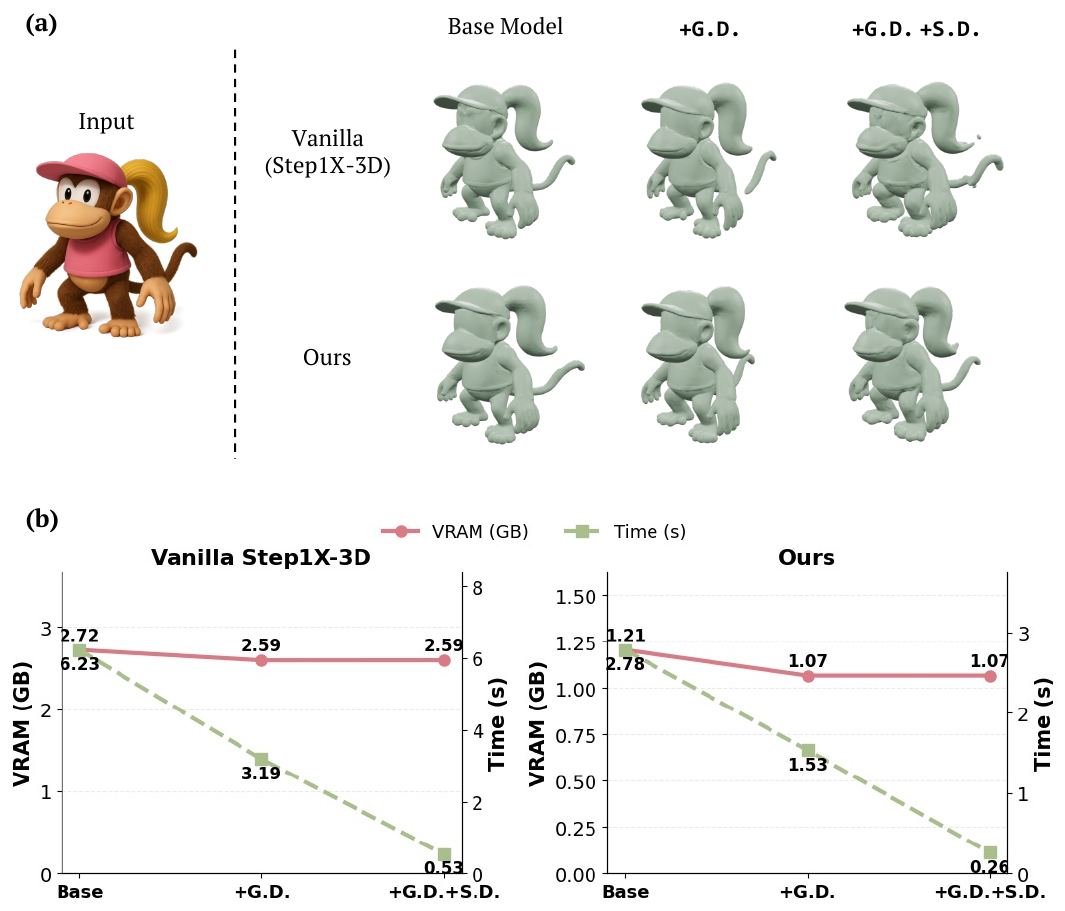}
    \caption{{%
        Additional results of secondary distillation for acceleration.
        \textbf{(a)} Qualitative comparisons.
        \textbf{(b)} Overhead comparisons.
    }}
    \label{fig:rebuttal_acc}
\end{figure}

We further compare against the original model equipped with guidance distillation and with both guidance and step distillation, as shown in \cref{fig:rebuttal_acc}.
The results indicate that compression can be combined with acceleration without introducing the degradation observed in the accelerated original model, while also reducing model footprint and inference cost.

\section{Detailed Analysis of Vitality Layers}
\label{sec:app_analysis}

\subsection{Analysis with Chamfer Distance Metrics}
\label{sec:app_analysis_cd}

\begin{figure}[t]
    \centering
    \includegraphics[trim={1mm 0mm 1mm 0mm}, clip, width=0.65\linewidth]{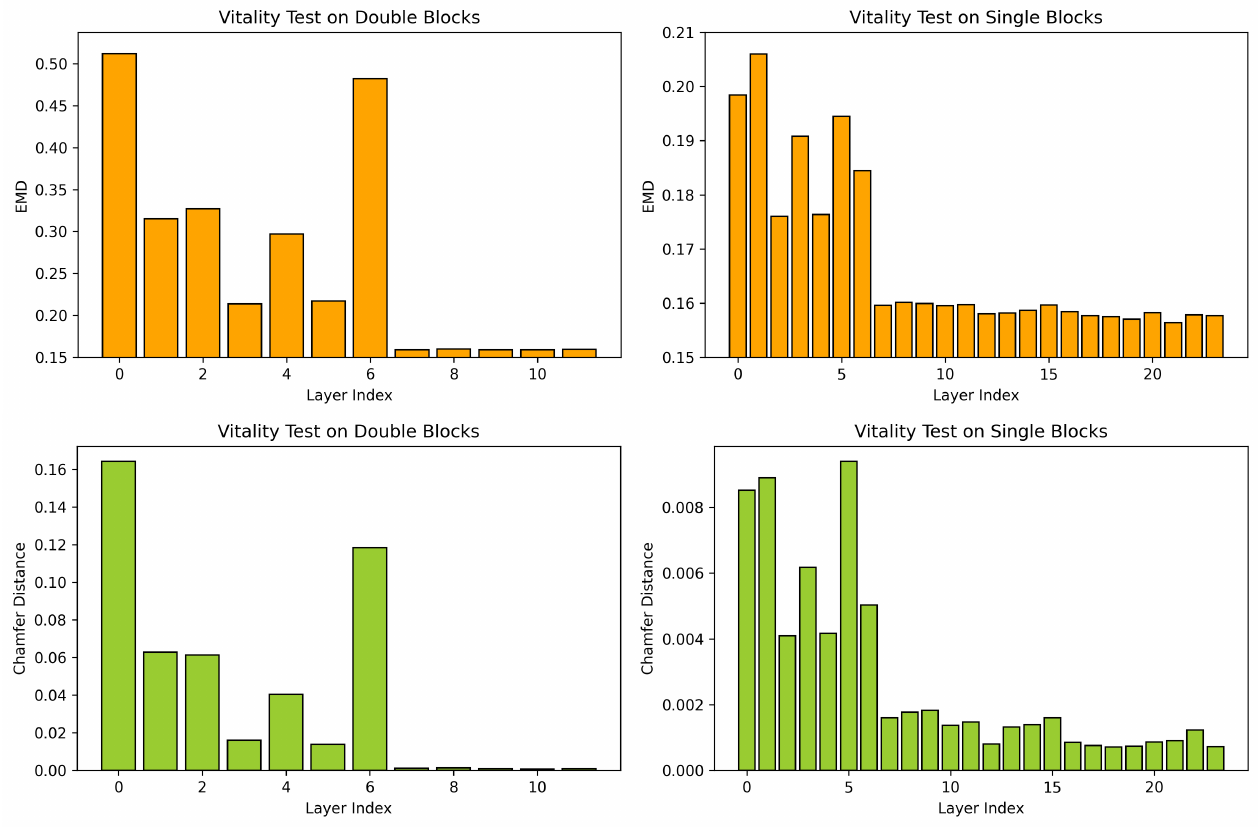}
    \caption{\textbf{Detailed Vitality Analysis of Step1X-3D.} Up: Vitality analysis result with Earth Mover's Distance (EMD). Down : Analysis result with Chamfer Distance.}
    \label{fig:step1x_chamfer}
\end{figure}

\begin{figure}[t]
    \centering
    \includegraphics[width=0.82\linewidth]{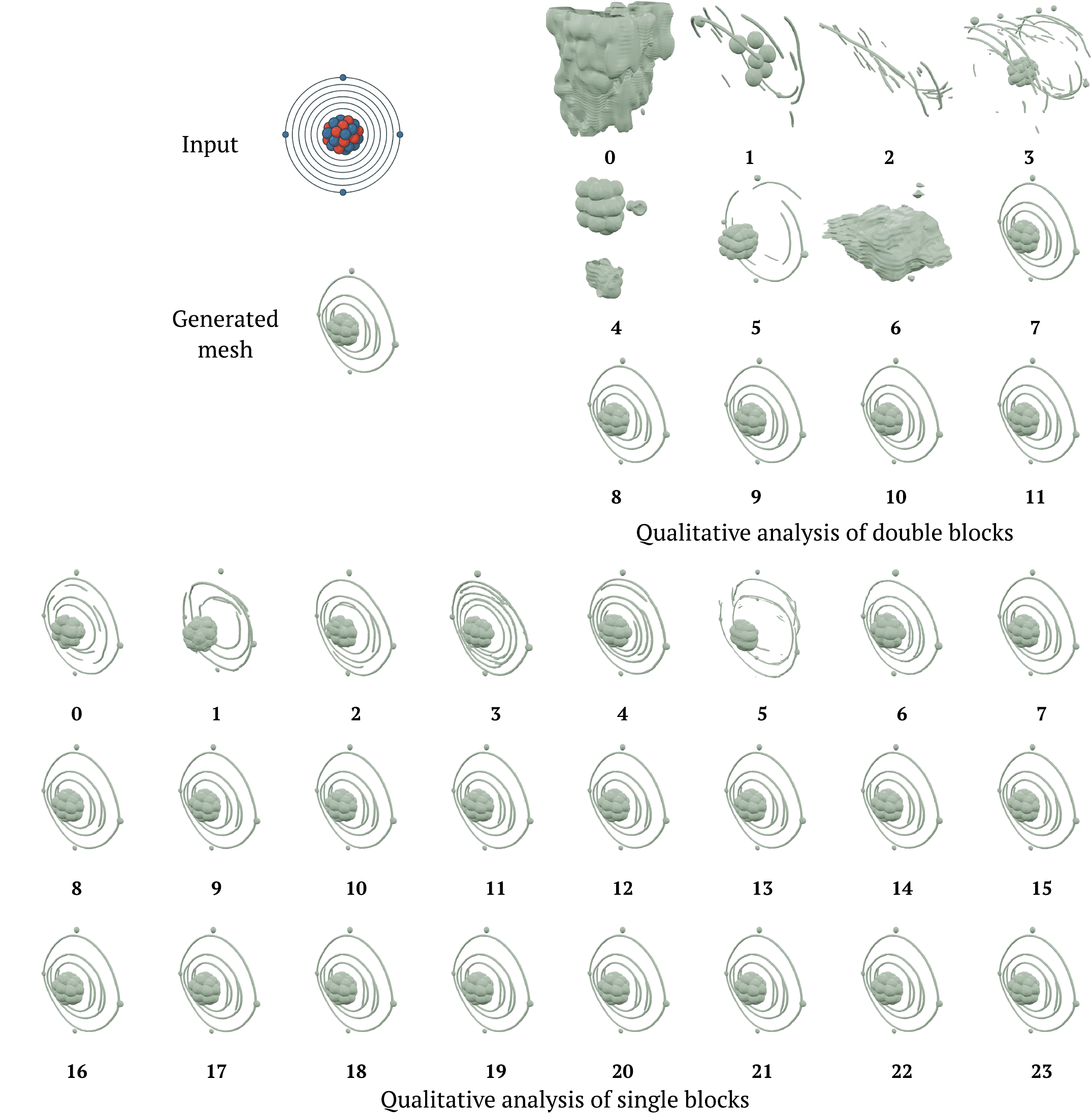}
    \caption{\textbf{Meshes Generated After Layer Removal (Step1X-3D).} Numbers below each mesh denote removed layer indices. Removing double block layer 0–6 or single block layer 0–6 significantly degrades quality (vital layers), while removing other layers (non-vital) has minimal effect.}
    \label{fig:step1x_anal_quali}
\end{figure}

To support the proposed vitality score calculation method, we further show the vitality score analysis on different distance metrics in \cref{fig:step1x_chamfer}. We show the analysis results of Chamfer distance.
The quantitative analysis mostly follow the analysis result using EMD. Again, the analysis of Chamfer distance also show clear difference of layer contribution to output image.
As shown in our analysis graph of \cref{fig:step1x_anal_quali}, we set non-vital layers as double block 7-11 and single block 7-23.
In the qualitative analysis results, we can still observe that changes in vital layers (single 0-6 , double 0-6) produce significant deformation or degradation of detailed structure, while changes in non-vital layers do not make any major difference.
The qualitative analysis again confirm our analysis results.

\subsection{Analysis on Hunyuan3D Models}
\label{sec:app_analysis_hunyuan}
\begin{figure}[t]
    \centering
    \includegraphics[width=0.65\linewidth]{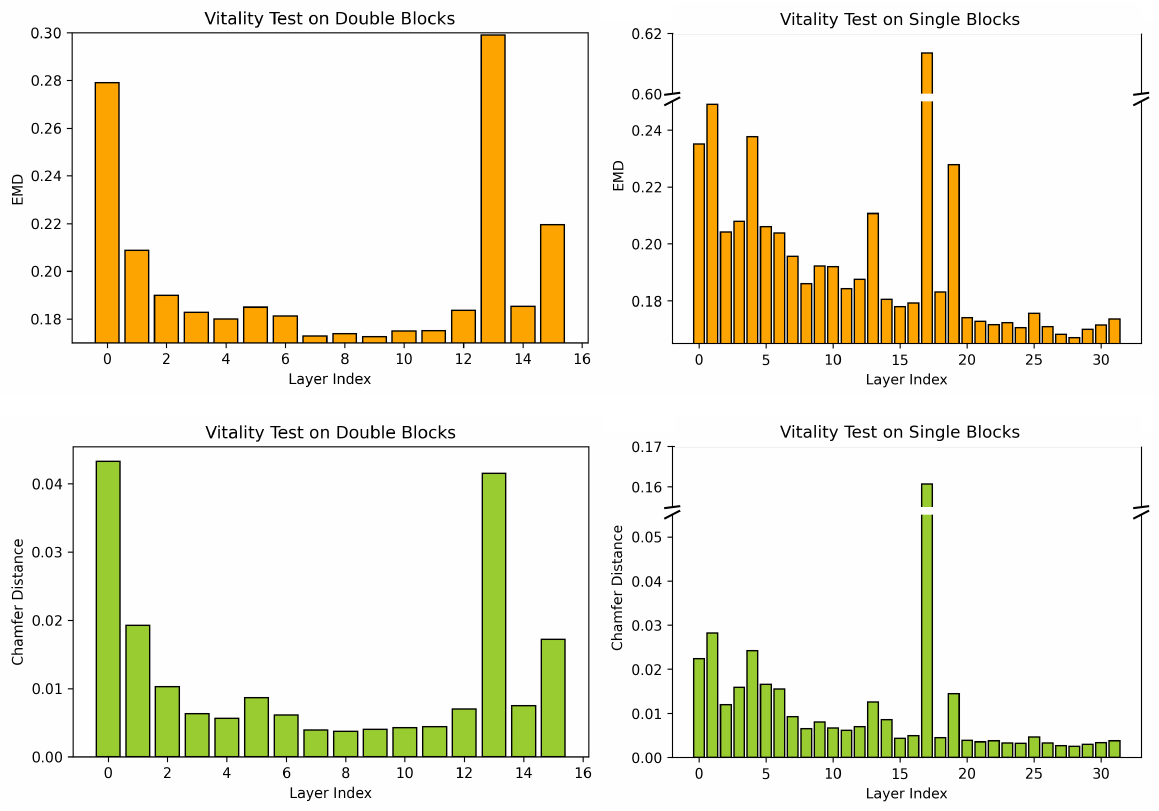}
    \caption{\textbf{Detailed Vitality Analysis of Hunyuan3D 2.0.} Up: Vitality analysis result with Earth Mover's Distance (EMD). Down : Analysis result with Chamfer Distance.}
    \label{fig:hy_chamfer}
\end{figure}

\begin{figure}[t]
    \centering
    \includegraphics[width=0.73\linewidth]{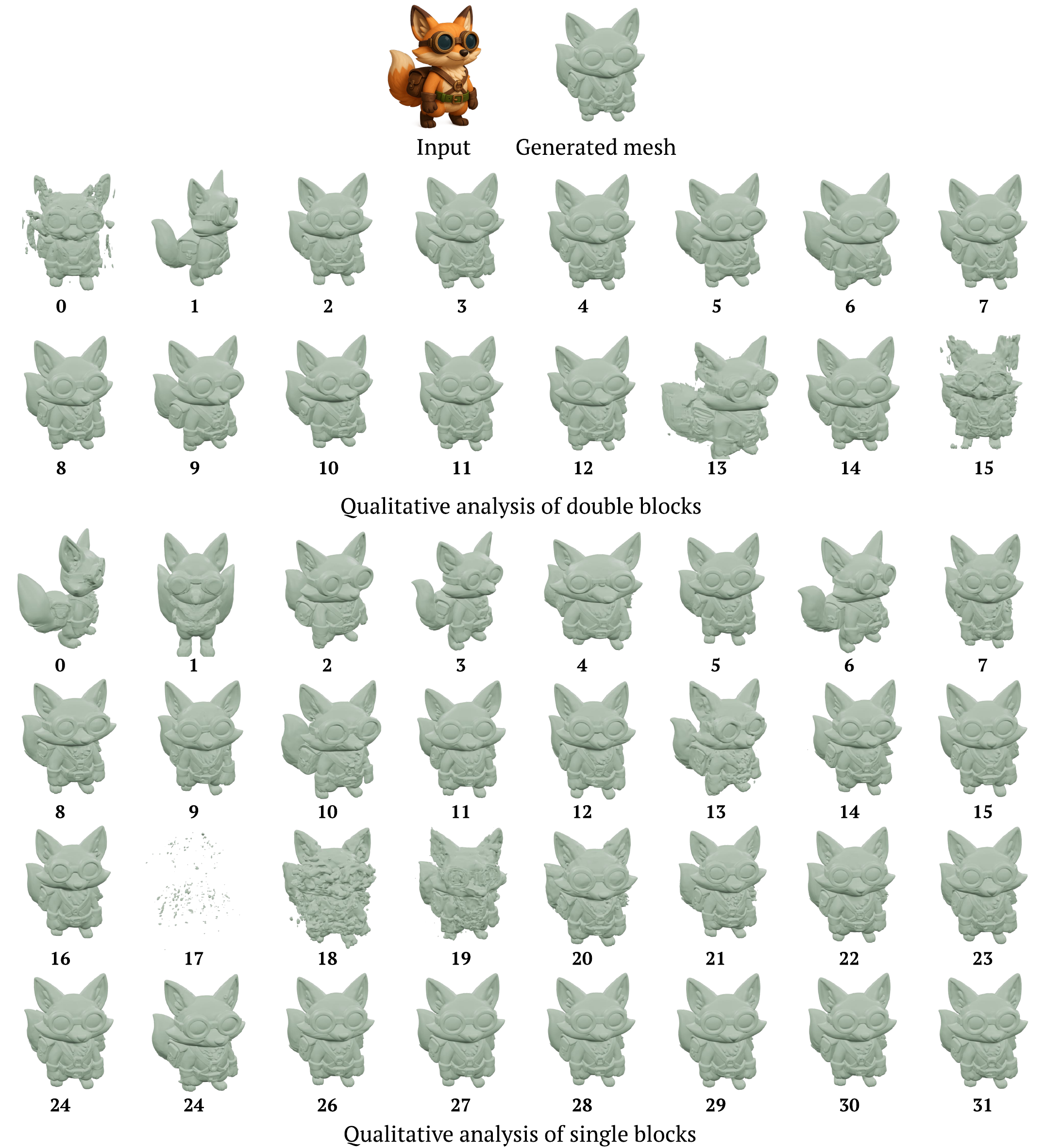}
    \caption{\textbf{Meshes Generated After Layer Removal (Hunyuan3D 2.0).} Numbers below each mesh denote removed layer indices. Removing certain vital layers leads to severe quality degradation. Especially, removing single block layer 17 results in the complete collapse of the mesh.}
    \label{fig:hy_anal_quali}
\end{figure}

We also conduct a layer analysis on Hunyuan3D 2.0 using our vitality score computation method in \cref{fig:hy_chamfer}.
Similar to Step1X-3D, we are able to distinguish between vital and non-vital layers; however, unlike Step1X-3D, where all layers beyond a certain index are non-vital, the Hunyuan model shows a mixed ordering of vital and non-vital layers.
Moreover, the difference between vital and non-vital layers is less pronounced compared to Step1X-3D.
This observation is also reflected in our ablation study: while Step1X-3D maintains performance close to the full model with layer pruning alone, the Hunyuan model exhibits slight artifacts without training.
In case of qualitative analysis in \cref{fig:hy_anal_quali}, modification of vital layers show severe deformation from original generated mesh as expected.
When we remove non-vital layers which has small distance, the output meshes still show slight difference in high-frequency details. 

\begin{figure}[t]
    \centering
    \includegraphics[width=0.65\linewidth]{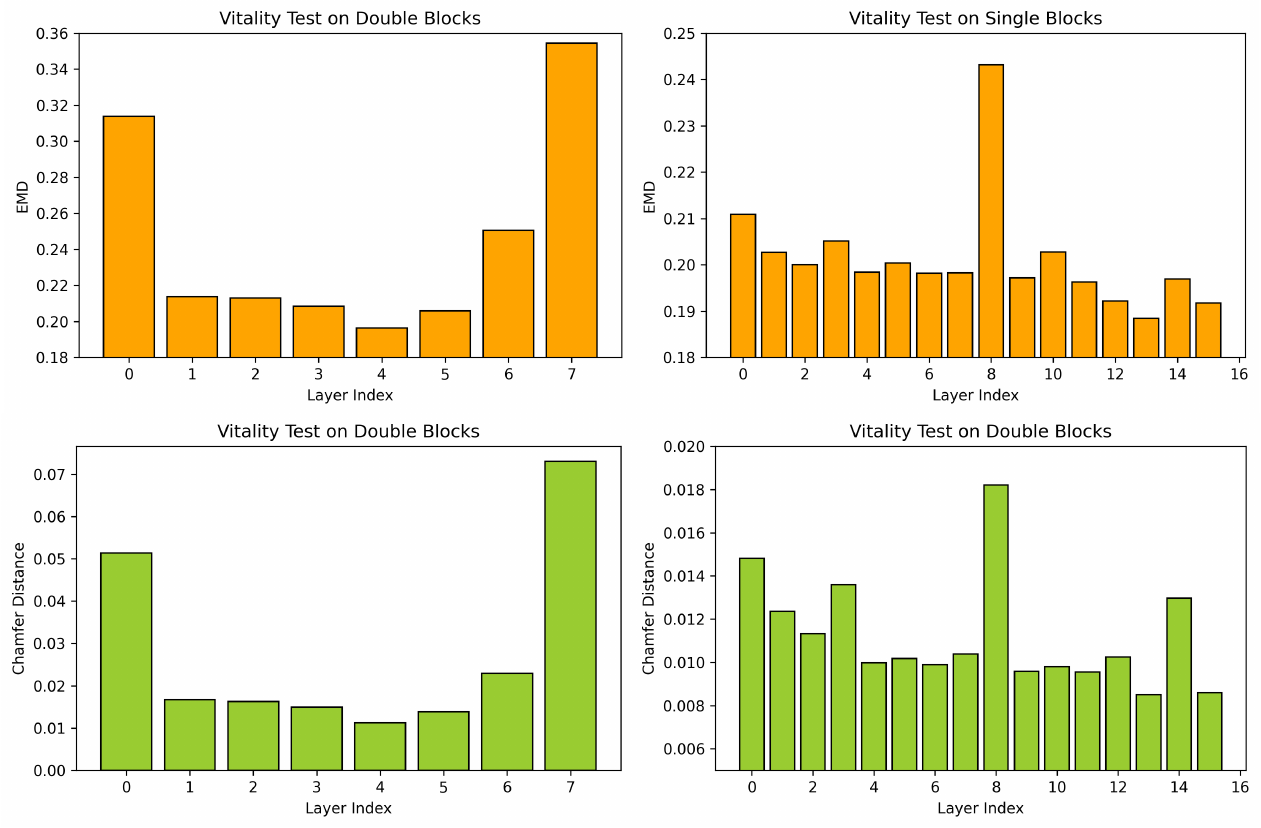}
    \caption{\textbf{Detailed Vitality Analysis of Hunyuan3D 2Mini.} Up: Vitality analysis result with Earth Mover's Distance (EMD). Down : Analysis result with Chamfer Distance.}
    \label{fig:hymini_chamfer}
\end{figure}

\begin{figure}[t]
    \centering
    \includegraphics[width=0.9\linewidth]{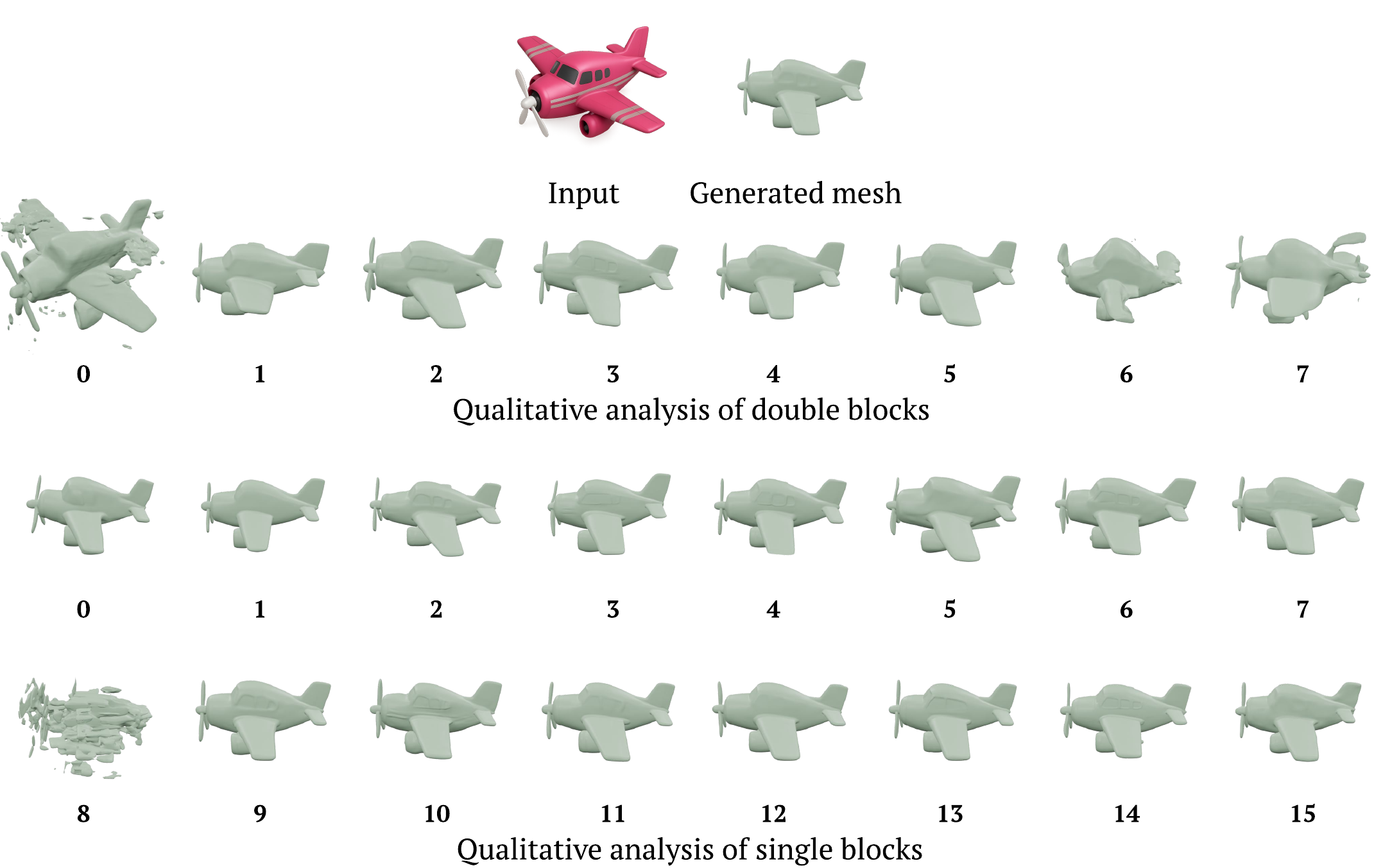}
    \caption{\textbf{Meshes Generated After Layer Removal (Hunyuan3D 2mini).} Numbers below each mesh denote removed layer indices. Similar to Hunyuan3D 2.0, removing certain vital layers (single block layer 8) results in severe quality degradation. Despite being a lightweight variant, the model still contains non-vital layers whose removal has little impact on performance.}
    \label{fig:hymini_anal_quali}
\end{figure}

For the Hunyuan3D 2mini model (\cref{fig:hymini_chamfer}), which is already a compressed model with significantly fewer layers than the original, our layer analysis reveals that the number of layers with low vitality score (which can be regarded as non-vital) is fewer compared to larger-scale models.
Consequently, the number of layers that can be pruned is more limited. Instead, we focus more on adaptive quantization with using used more 4-bit layers.
In our qualitative analysis in \cref{fig:hymini_anal_quali}, we can see that when removing the double layers, all the mesh outputs show geometric deformation from original meshes.
In single block layers, we can also see there are some level of deformation in mesh details when removing front layers (0-13).

\subsection{Vitality Analysis on TRELLIS}
\label{sec:vitality_trellis}

\begin{figure}[t]
    \centering
    \includegraphics[width=0.95\linewidth]{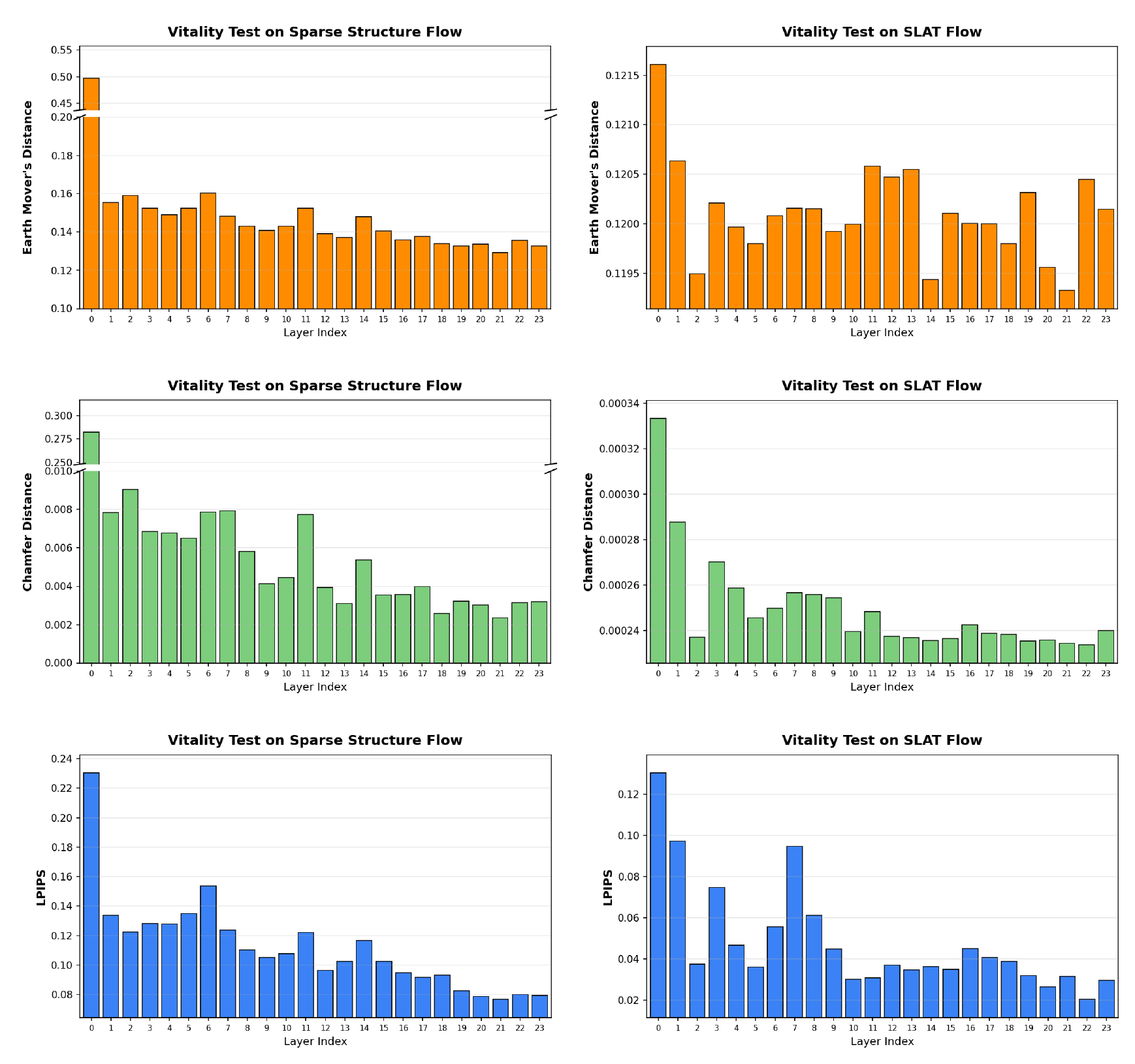}
    \caption{\textbf{Detailed Vitality Analysis of TRELLIS.}
    Up: Vitality analysis result with Earth Mover's Distance (EMD).
    Middle: Analysis result with Chamfer Distance.
    Down: Analysis result with LPIPS.
    }
    \label{fig:trellis_plot}
\end{figure}

TRELLIS~\cite{xiang2024structured} is a state-of-the-art 3D generative model that, like Step1X-3D and Hunyuan3D 2.0, employs transformer-based flow models as its generative backbone. It therefore lends itself naturally to the same layer-wise vitality analysis. However, a key architectural distinction sets TRELLIS apart from the two aforementioned models: whereas Step1X-3D and Hunyuan3D 2.0 decouple geometry and texture into independent generation stages, TRELLIS encodes both structural and textural information jointly within its Structured LATent (SLAT) representation. Specifically, TRELLIS adopts a two-stage pipeline consisting of a Sparse Structure Flow that predicts which voxels are active, followed by a SLAT Flow that generates local latent vectors---encoding both geometry and appearance---for each active voxel. Because geometry and texture are entangled in a single latent space, evaluating the impact of layer removal requires assessing not only geometric fidelity but also visual quality. We therefore measure EMD and CD for geometry, and additionally compute LPIPS between rendered images of the textured output meshes. The full results are presented in Fig.~\ref{fig:trellis_plot}.

\paragraph{Sparse Structure Flow.}
Among all layers, layer~0 exerts the strongest influence, producing the largest degradation across all three metrics. Beyond layer~0, layer~6 emerges as the next most impactful, particularly in LPIPS. Notably, when excluding layer~0, the EMD values remain below 0.16 and CD values stay under 0.01, suggesting that no single layer in this stage is catastrophically vital to geometric quality on its own. Texture quality, however, tells a different story: LPIPS reveals that earlier layers (including layers~0 and~6) contribute substantially to appearance fidelity, while later layers (20--23) can be removed with relatively minor visual degradation.

\begin{figure}[t]
    \centering
    \includegraphics[width=0.95\linewidth]{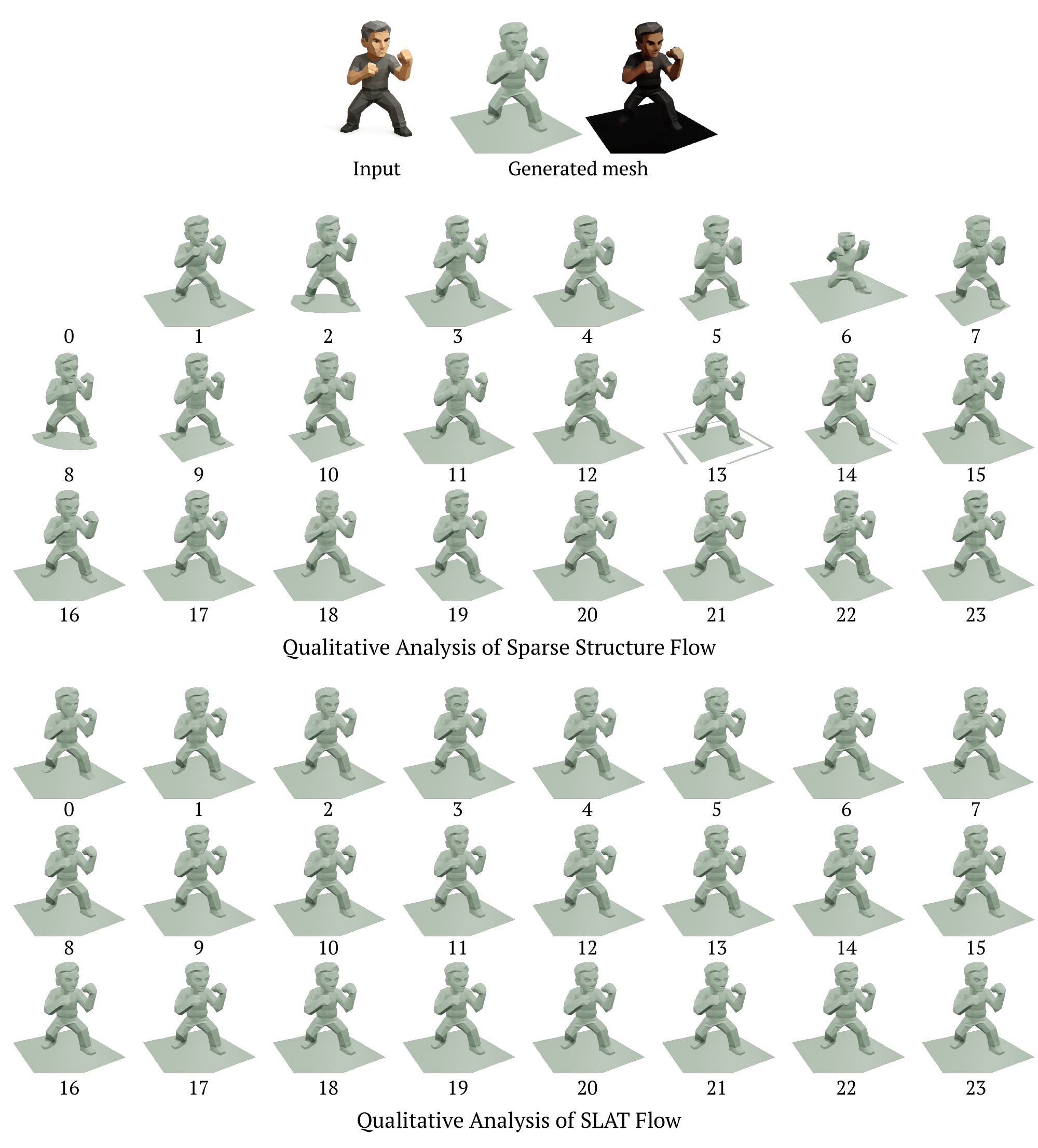}
    \caption{\textbf{Meshes (Only Geometry) Generated After Layer Removal (TRELLIS).}
    Numbers below each mesh denote removed layer indices.}
    \label{fig:trellis_render_geo}
\end{figure}

\begin{figure}[t]
    \centering
    \includegraphics[width=0.95\linewidth]{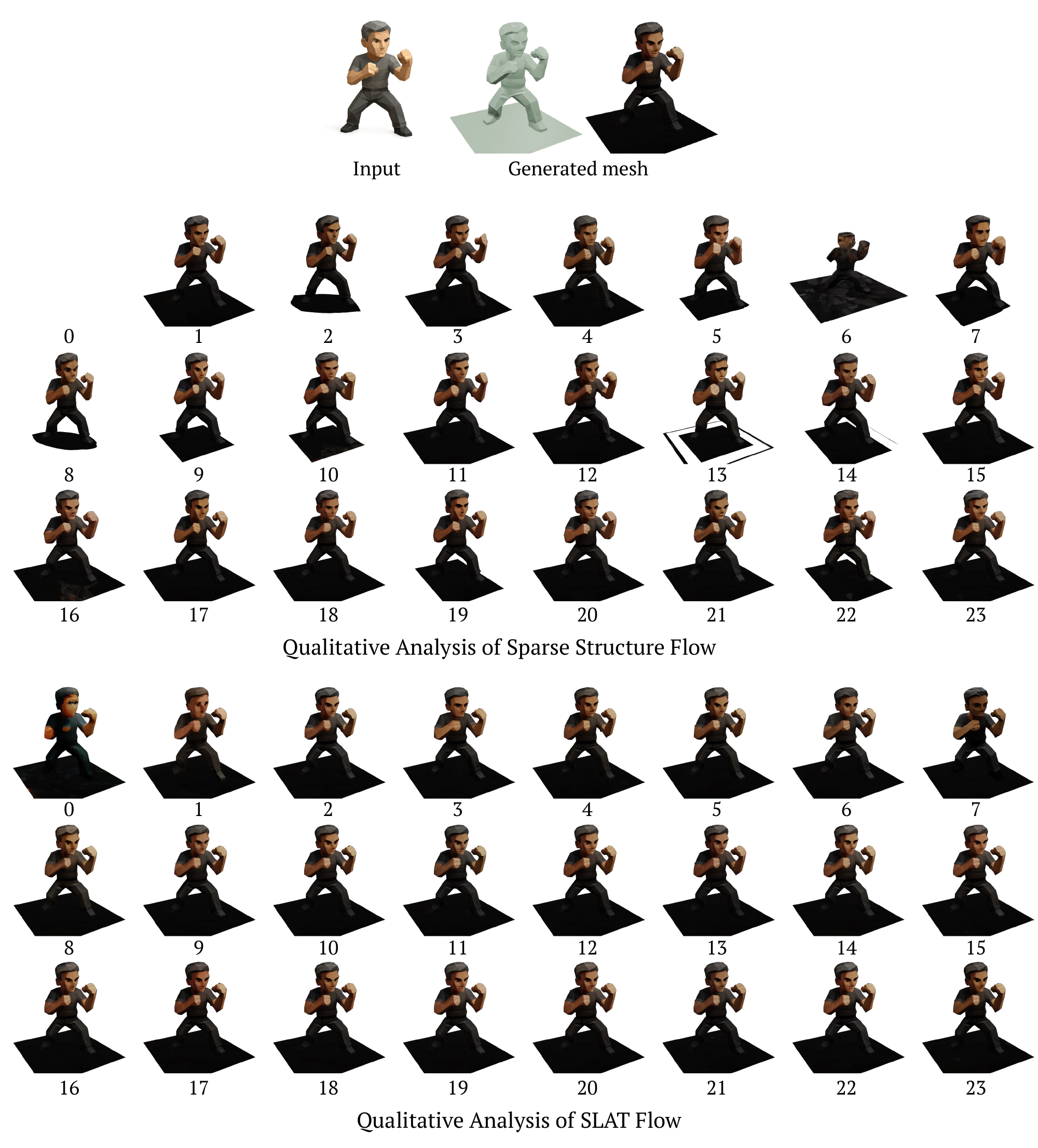}
    \caption{\textbf{Meshes (Textured) Generated After Layer Removal (TRELLIS).}
    Numbers below each mesh denote removed layer indices.}
    \label{fig:trellis_render_tex}
\end{figure}

\paragraph{SLAT Flow.}
The SLAT Flow exhibits a markedly different vitality profile. Its impact on geometry is minimal: the difference between the maximum and minimum EMD across all layers is approximately 0.002, and the corresponding CD variation is on the order of $1 \times 10^{-4}$. This indicates that individual SLAT Flow layers contribute negligibly to geometric quality, which is expected given that the coarse 3D structure has already been established by the Sparse Structure Flow. Visual inspection of the geometry in Fig.~\ref{fig:trellis_render_geo} corroborates this finding. In contrast, LPIPS measurements reveal pronounced vitality for texture quality. As illustrated in Fig.~\ref{fig:trellis_render_tex}, removing layers~0, 1, or~7 from the SLAT Flow leads to noticeable texture degradation compared to the baseline, confirming that these layers play a critical role in appearance synthesis.

\paragraph{Implications For Compression.}
Overall, the vitality analysis shows that TRELLIS is amenable to layer-wise
compression, while also revealing the limitation of relying solely on geometric
metrics. In the Sparse Structure Flow, which primarily governs geometry
generation, EMD-based vitality serves as an effective criterion for compression,
as supported by our Sparse Structure Flow compression results in Sec.~\ref{sec:trellis_applicability}~\refofpaper{}. In contrast, the SLAT Flow primarily affects texture and appearance generation: EMD and CD remain
nearly unchanged across layer removals, whereas LPIPS varies substantially and
identifies layers critical to visual fidelity. These results suggest that
compressing TRELLIS beyond the Sparse Structure Flow requires a joint criterion
that accounts for both geometry and appearance. We therefore leave
LPIPS-guided compression of the SLAT Flow, and the resulting end-to-end
compression of the full TRELLIS architecture, as future work.

\section{Limitations and Future Work}
\label{sec:app_limitations}

\begin{figure}[t]
    \centering
    \includegraphics[width=0.9\linewidth]{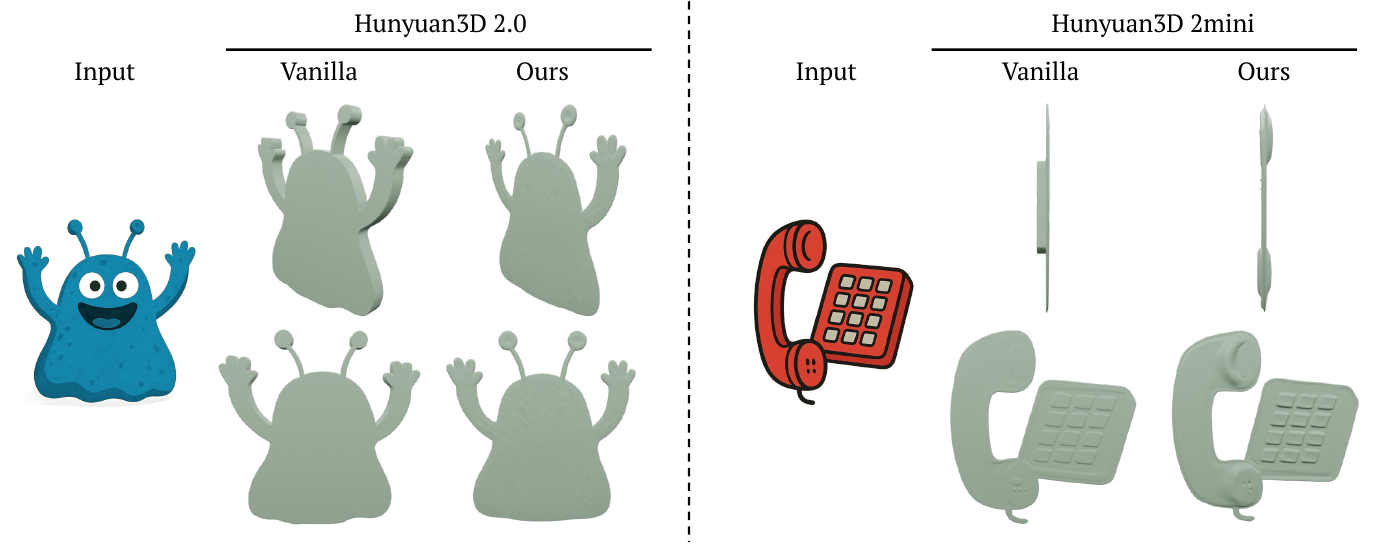}
    \caption{
    \textbf{Limitations.}
    Since our approach aims for model compression while ``maintaining'' performance, the compressed model still shares core geometric limitations of the original framework.
    }
    \label{fig:app_failures}
\end{figure}

As mentioned in the main paper, our method compresses 3D DiT models by up to 66\% while maintaining performance comparable to the full-parameter model.
Although the proposed quantization framework supports precision down to 4 bits, we did not explore more extreme settings (\eg, 1–2 bit), which would require dedicated hardware-level implementations.
Moreover, the compressed model inherits the inherent geometric and topological limitations of the original architecture.
As illustrated in \cref{fig:app_failures}, Hunyuan-based models still struggle to reconstruct accurate 3D structures from flat or stylized illustrations, since our distillation-based compression is designed to preserve the behavior of the original model.

For future work, we plan to further accelerate inference by reducing sampling steps and distilling away classifier-free guidance.
We also aim to extend the framework to texture generation models, enabling efficient joint optimization of both geometry and texture generation.
Finally, since the current vitality thresholds are manually tuned for each architecture, we plan to automate this process using relative vitality statistics across architectures to improve general applicability while maintaining the plug-and-play nature of our approach.

\newpage
\clearpage



\newpage
\clearpage


\end{document}